\documentclass[10pt,twocolumn,letterpaper]{article}

\usepackage[pagenumbers]{cvpr} 

\usepackage{microtype}

\definecolor{cvprblue}{rgb}{0.21,0.49,0.74}
\usepackage[pagebackref,breaklinks,colorlinks,allcolors=cvprblue]{hyperref}

\usepackage{placeins}
\usepackage{etoolbox}

\usepackage{graphicx}
\usepackage{makecell}
\usepackage[table]{xcolor}

\usepackage{booktabs}
\usepackage{pifont}
\usepackage{enumitem}
\newcommand{\cmark}{\ding{51}}   
\newcommand{\xmark}{\ding{55}}   
\newcommand{\pmark}{$\circ$}     

\def\confName{CVPR}
\def\confYear{2027}

\title{SLUSH: Spatially Consistent 3D Lung Shape Reconstruction\\from Unaligned 2D Clinical X-Rays}

\author{Martin Rath\textsuperscript{1,2}\thanks{Corresponding author.}, Fabian Bongratz\textsuperscript{1,2}, Christian Wachinger\textsuperscript{1,2}\\
\textsuperscript{1}Technical University of Munich (TUM), Munich, Germany\\
\textsuperscript{2}Munich Center for Machine Learning (MCML), Munich, Germany\\
{\tt\small martin.rath@tum.de}
}

\begin{document}
\maketitle
\begin{abstract}
Reconstructing 3D anatomical structures from routine 2D X-rays has high potential to improve access to precise volumetric and shape-based assessments for treatment planning and diagnosis. However, real-world translation of existing methods is constrained by the widespread use of synthetic projections for development and testing. Critically, such synthetic datasets are perfectly aligned by construction, whereas real-world radiographs exhibit substantial variations in acquisition geometry due to imperfect positioning and imaging procedures. To address this gap, we propose SLUSH (Spatially consistent LUng SHape reconstruction), a framework for 3D shape reconstruction from unaligned bi-planar X-rays. A dedicated pose-estimation head predicts the acquisition geometry (i.e., the extrinsic X-ray parameters) before performing template-based reconstruction guided by deep image features. In contrast to prior work, SLUSH is trained on imperfect, separately acquired X-ray--CT pairs, reflecting routine clinical acquisition conditions and enabling scalable training on routinely acquired retrospective data. To enable evaluation under realistic acquisition variability, we introduce LIDC-RealPairs, a processed subset of the LIDC-IDRI dataset with real bi-planar X-ray--CT pairs. On bilateral lung shape reconstruction, SLUSH significantly improves over the state of the art, reducing surface reconstruction errors by 30.6\% on LIDC-RealPairs and by 20.7\% on an independent in-house dataset. We further demonstrate the utility of the reconstructed shapes for downstream 3D nodule localization using only bi-planar radiographs as input. Code and benchmark is available at: https://github.com/ai-med/SLUSH.
\end{abstract}
    
\section{Introduction}
\label{sec:intro}


Chest radiography is the most commonly performed radiographic examination, accounting for nearly half of all diagnostic radiographic procedures in the United States~\cite{mettler2009radiologic}. Standard clinical acquisition consists of posteroanterior (PA) and lateral projections~\cite{american2017acr}. Owing to its low cost~\cite{hakkaart-van_roijen_costing_2024}, low radiation exposure~\cite{mettler2008effective}, and broad availability, chest radiography serves as the first-line imaging modality for thoracic assessment~\cite{batra2025acr}. However, information obtained from radiographs is largely limited to two-dimensional projections and coarse scalar measurements, such as lung height for transplant matching~\cite{guillamet2022uncertainty} or lung volume estimates~\cite{sogancioglu2022automated, ghimire2024estimating}. Furthermore, the projective nature of X-rays obscures the three-dimensional location of anatomical structures and pathologies, such as lesions, limiting their utility for tasks that require precise spatial localization, including operative planning.

\begin{figure}[t]
    \centering
    \includegraphics[width=\linewidth]{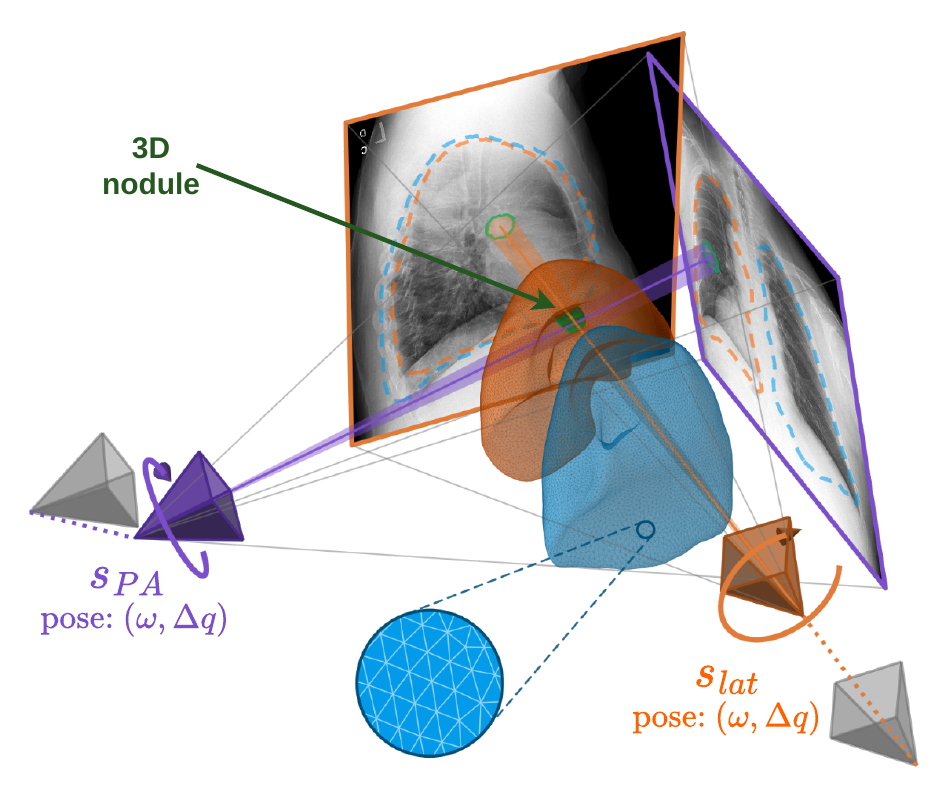}
    \caption{SLUSH reconstructs 3D lungs from real X-rays with an unknown geometry. A pose head corrects each camera (gray) by $\omega$ and $\Delta q$ (colored), so the lungs project consistently onto both views. An annotated nodule is back-projected through these cameras and localized in 3D.}
    \label{fig:teaser}
\end{figure}

For three-dimensional assessment of thoracic anatomy, CT is often the optimal clinical imaging modality. However, CT is typically reserved for specific indications, such as an inconclusive radiograph or suspected complications~\cite{batra2025acr}, and remains scarce in many low- and middle-income countries~\cite{frija2021improve, world2017global}. Recently, methods have emerged that aim to recover three-dimensional information directly from routine chest radiographs, thereby circumventing additional radiation exposure or acquisition costs. Existing approaches pursue this goal by synthesizing CT volumes~\cite{ying2019x2ct, kyung_perspective_2023}, reconstructing voxel grids~\cite{PAN2026104085, liu2025swin}, predicting implicit shape representations~\cite{guven2026x2br}, or deforming anatomical templates~\cite{wang2019deeporgannet, nakao2022igcn}.

However, these methods rely on digitally reconstructed radiographs (DRRs) for both training and evaluation. DRRs are synthetically generated from CT scans with perfectly aligned geometry, producing image pairs that do not reflect real-world clinical practice. In routine care, CT and radiographic examinations are acquired independently and often differ in patient orientation, pose, and respiratory state. These discrepancies can be substantial: in our in-house data, we observed average misalignments of $10^\circ$ in rotation and 16\,cm in translation. Beyond geometry, real radiographs differ from DRRs due to acquisition- and patient-specific effects, including scatter, detector response, and post-processing; the combination of these factors cannot be faithfully simulated by simple augmentations. Therefore, the real-world generalization of methods developed on synthetic DRRs is severely limited.


To address this gap, we propose SLUSH, a 3D shape reconstruction framework for real-world radiographs. Unlike previous approaches, SLUSH estimates the unknown acquisition geometry of each view with a pose-estimation head, which allows it to be trained and evaluated on bi-planar X-rays paired with separately acquired CT scans rather than on perfectly aligned synthetic data. To support reproducible research and benchmarking under real-world acquisition conditions, we release \emph{LIDC-RealPairs}, our preprocessed version of the public LIDC-IDRI dataset~\cite{armato_iii_lung_2011} comprising 179 real X-ray--CT pairs (\cref{sec:benchmark}). We further evaluate the generalization of SLUSH on retrospective in-house clinical data and demonstrate its ability to localize structural abnormalities in 3D directly from bi-planar radiographs. An example of pulmonary nodule localization is shown in \cref{fig:teaser}.





\noindent In summary, our key contributions are:
\begin{itemize} 
\item To our knowledge, this is the first work to train and evaluate 3D lung reconstruction on clinical bi-planar X-rays paired with separately acquired CT scans, rather than on simulated projections or co-registered CT scout images. 
\item SLUSH, a template-based mesh reconstruction model that explicitly disentangles pose from shape. Its pose head learns the acquisition geometry largely from the reconstruction losses, which allows training directly on unaligned real X-ray--CT pairs.
\item LIDC-RealPairs, the first public benchmark for 3D shape reconstruction from clinical bi-planar X-rays with DICOM geometry paired with separately acquired CT scans. 
\item A comprehensive evaluation on clinically relevant downstream tasks, including lung volumetry competitive with dedicated regression methods and 2D-to-3D pulmonary nodule localization, demonstrating significant improvements over the state of the art.
\end{itemize}


\section{Related work}
\label{sec:related_work}

Conditional reconstruction models recover 3D shape from a small number of calibrated views by decoding pixel-aligned image features into implicit functions \cite{saito2019pifu}, radiance fields \cite{yu2021pixelnerf}, or deformed meshes \cite{wang2018pixel2mesh, wen2019pixel2mesh++}. More recently, large generative models \cite{huang2025spar3d, xu2024instantmesh, zhao2025hunyuan3d, xiang2025structured} have enabled feed-forward reconstruction of closed object meshes directly from images. These methods are trained on large collections of synthetic 3D assets and do not require calibrated cameras or metric scale. Point-map estimators \cite{wang2024dust3r, wang2025vggt} instead predict a 3D point for each image pixel under the assumption that every viewing ray intersects a single surface. 

These assumptions break down for radiographs. X-rays integrate attenuation along each ray, while the lungs are internal structures that overlap surrounding anatomy in projection. As a result, only the appearance-independent feature projection paradigm of conditional reconstruction models readily transfers to this setting. Existing X-ray reconstruction approaches either synthesize CT volumes from \mbox{single-,} bi-planar, or multi-view radiographs \cite{ying2019x2ct, kyung_perspective_2023, cahan2025x, rath2026axon, liu2024diffux2ct}, which typically requires a subsequent segmentation step, or reconstruct the anatomy directly as a voxel mask \cite{PAN2026104085, liu2025swin, kasten2020end}, implicit field \cite{guven2026x2br, yu2025spider}, deformed template \cite{wang2019deeporgannet, nakao2022igcn, shi2024xraywizard}, or statistical shape model \cite{bhalodia2018deepssm}. \cref{tab:paradigms} compares the different properties of these paradigms. A major challenge is the lack of publicly available paired real bi-planar X-ray and CT datasets. Consequently, most methods are trained on digitally reconstructed radiographs (DRRs) \cite{galvin1995use}. The appearance gap between DRRs and clinical radiographs has been widely studied \cite{gao2023synthetic, unberath2018deepdrr} and can be partially mitigated by training on real X-rays \cite{rath2026axon}. In contrast, the geometric gap has received comparatively little attention. Because DRRs are rendered directly from the target CT volume, the input radiographs and reconstruction target share the same patient pose, breathing state, and a perfectly known, typically orthogonal, projection geometry. Clinical radiographs differ substantially: they are acquired in an upright position, often days apart from the supine CT scan, possibly at a different level of inspiration, and with only approximately known source distances. Moreover, PA and lateral views are acquired sequentially rather than simultaneously. Posture changes alone increase total lung volume by approximately $11\%$ between supine and standing acquisition \cite{yamada2020differences}. The few methods trained on real radiographs either do not explicitly model acquisition geometry \cite{cahan2025x, rath2026axon} or rely on CT scout images that are largely co-registered with the target CT \cite{shi2024xraywizard}. None explicitly addresses the resulting misalignment between input images and reconstruction targets (\cref{tab:paradigms}, Pose). 2D/3D registration methods can recover projection geometry using differentiable rendering \cite{gopalakrishnan2022fast, gopalakrishnan2024diffpose}, patient-specific neural networks \cite{gopalakrishnan2026rapid}, or deformable models \cite{gopalakrishnan2026polypose}. However, these approaches require access to the patient's CT volume and can therefore align training pairs but not estimate acquisition geometry at test time. In contrast, we learn to canonicalize projection geometry directly within the reconstruction model, without requiring CT data during inference. 

A related line of work estimates lung volume directly from radiographs. Early methods relied on geometric measurements \cite{barnhard1960roentgenographic}, while more recent deep learning approaches regress total lung volume from single-view or bi-planar X-rays \cite{sogancioglu2022automated, ghimire2024estimating}, including one method trained on real bi-planar acquisitions \cite{sogancioglu2022automated}. Other approaches estimate thickness maps and integrate them to obtain lung volume \cite{dorosti2025estimating}. In lung transplantation, size matching is still commonly based on simple linear measurements such as lung height \cite{guillamet2022uncertainty}. Unlike our approach, these methods provide scalar or linear measurements rather than patient-specific 3D shape information. We compare against dedicated volume-regression methods in \cref{sec:comparison_related_work}.

\begin{table}[t]
\caption{Properties of reconstruction paradigms.
\textbf{Vol.}: direct volumetry from the output.
\textbf{Corr.}: consistent vertex correspondence across subjects.
\textbf{Topo.}: closed, watertight surface guaranteed.
\textbf{Pose}: explicit separation of pose from shape.
\textbf{Expr.}: shapes not restricted to a linear subspace.
\cmark~yes, \pmark~partial (CT synthesis requires additional segmentation,
implicit fields require surface extraction, SSMs are limited to the span of their principal modes), \xmark~no.}
\centering
\small
\setlength{\tabcolsep}{4pt}
\begin{tabular}{lccccc}
\toprule
Paradigm & Vol. & Corr. & Topo. & Pose & Expr. \\
\midrule
CT synthesis         & \pmark & \xmark & \xmark & \xmark & \cmark \\
Voxel segmentation   & \cmark & \xmark & \xmark & \xmark & \cmark \\
Implicit fields      & \pmark & \xmark & \xmark & \xmark & \cmark \\
SSM regression       & \cmark & \cmark & \cmark & \xmark & \pmark \\
Template deformation & \cmark & \cmark & \cmark & \xmark & \cmark \\
\midrule
Ours                 & \cmark & \cmark & \cmark & \cmark & \cmark \\
\bottomrule
\end{tabular}

\label{tab:paradigms}
\end{table}

\section{Method}
\label{sec:method}



\paragraph{Problem setting.}
Given a PA X-ray $x_{\text{PA}}$ and a left lateral X-ray $x_{\text{lat}}$ of the same patient, we aim to reconstruct surface meshes of the left and right lung, $\mathcal{M} = \{\mathcal{M}_\text{L}, \mathcal{M}_\text{R}\}$. Each X-ray is a perspective projection described by a camera with intrinsic and extrinsic parameters. The \emph{intrinsics} characterize the imaging system and are stored in the DICOM header as the \emph{Source-to-Detector Distance} (SDD) and the \emph{Imager Pixel Spacing} (IPS). The \emph{extrinsics} describe how the imaging system is positioned relative to the patient. Since source and detector move as one rigid unit, they consist of six parameters per view, three for rotation and three for translation. These are not recorded, so the acquisition angle, the \emph{Source-to-Object Distance} (SOD), and the in-plane position of the patient on the detector are unknown. Without them, a pixel cannot be related to a location in 3D.

As no external coordinate system is available, we express the extrinsics relative to the patient's anatomy in a \emph{canonical frame}, whose origin lies at the center of the lungs and whose axes follow the anatomical directions. The template mesh, a mean lung shape, is defined in this frame. The task is then to estimate the camera pose of each view in the canonical frame and the lung shape in the same frame. 
A CT of the same patient is available only for training and evaluation. It provides the ground-truth lung contours in 3D and, through 2D/3D registration with the X-rays, reference camera poses. At inference, the model receives only the two X-rays and their intrinsics, which mirrors the clinical practice.

\begin{figure*}
    \centering
    \includegraphics[width=\textwidth]{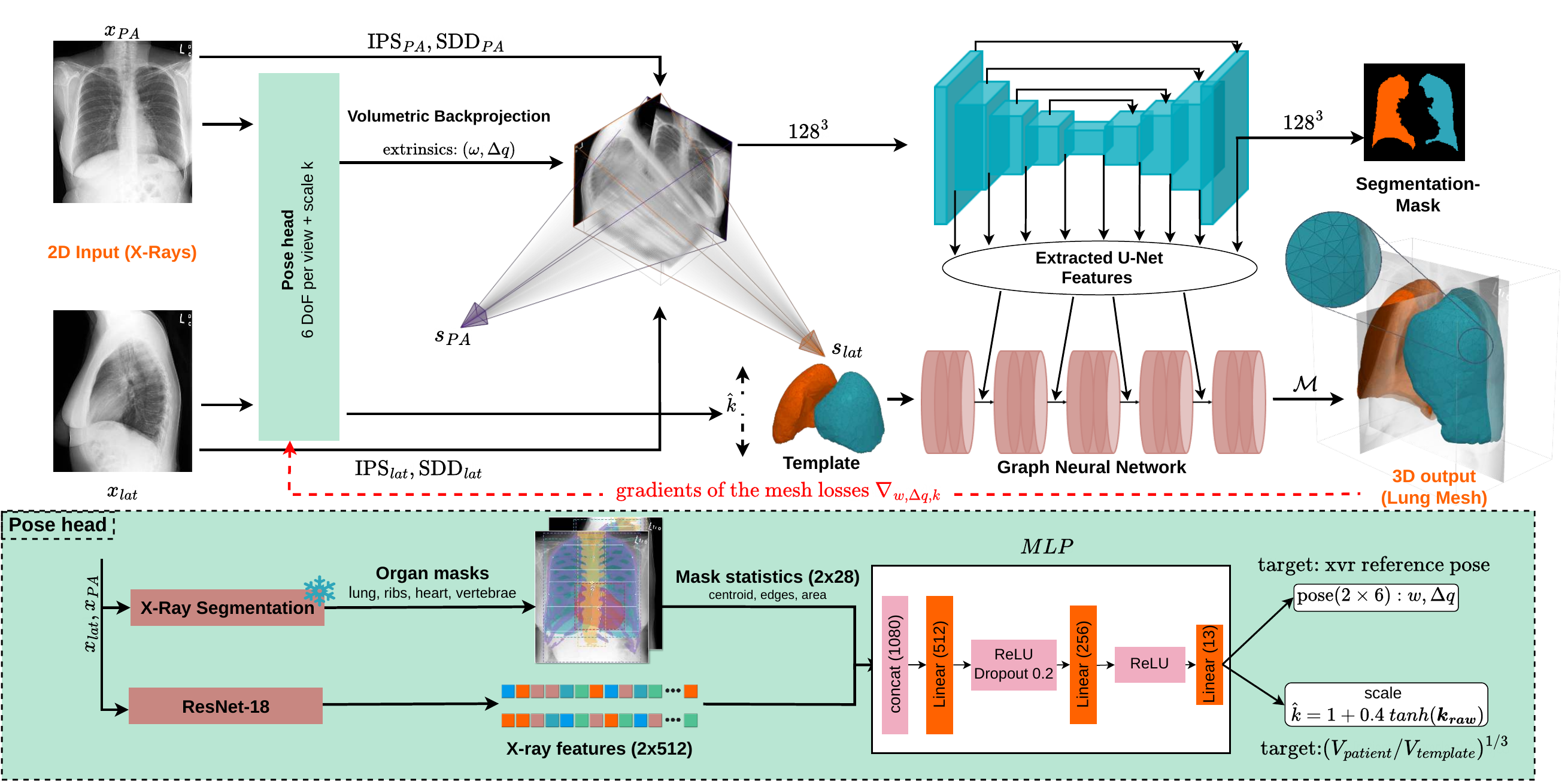}
    \caption{Method overview. \textbf{Top:} The pose head predicts the camera pose $(\omega,\Delta q)$ of each view and a template scale $k$. Using these poses and the camera intrinsics, both X-rays are back-projected into a $128^3$ volume, which the U-Net uses for reconstructing the lung in voxel space. With the help of the U-Net features, the GNN deforms the template, scaled by $k$, into the lung meshes $\mathcal{M}$. The back-projection is differentiable, so the mesh losses also train the pose head (dashed line). \textbf{Bottom:} The pose head combines mask geometry from an X-ray segmentation model with ResNet-18 image features and maps them with an MLP to the poses and $k$.}
    \label{fig:architecture}
\end{figure*}

\paragraph{Overview.}
SLUSH addresses this reconstruction problem in three steps, separating pose from shape to avoid entangling anatomical variation with acquisition geometry. See \cref{fig:architecture} for an overview. 
First, a \emph{pose head}, implemented by a neural network that takes both X-rays as input, predicts the extrinsics of each view as a deviation from a standard PA or lateral camera, together with a global scale $k$ that adapts the template to the patient's lung size (\cref{sec:pose_head}). Second, both X-rays are \emph{back-projected} into the canonical frame. Every voxel is assigned the intensities of the pixels it projects to under the predicted cameras, so that corresponding structures of the two views meet at the same 3D location (\cref{sec:diff_back}). Third, a 3D U-Net extracts features from the resulting backprojected volume. Those features are then projected onto the template vertices and processed by a graph neural network (GNN) to obtain a vertex-wise deformation field that transforms the template into 3D lung contours (\cref{sec:template_deformation}). 
Since the back-projection operation is differentiable, the pose head can be optimized not only with pose supervision from 2D/3D registration but also with mesh reconstruction losses. Note that the registration-derived reference poses are required only during training.

\subsection{Pose head}
\label{sec:pose_head}

\paragraph{Extrinsic references.}
Since the true extrinsics are not recorded, we estimate reference poses with xvr~\cite{gopalakrishnan2026rapid}, a 2D/3D registration method that searches for the camera pose under which a DRR rendered from the patient's CT best matches the real X-ray. It first trains a patient-specific network on DRRs with known poses to obtain an initial estimate, and then refines this estimate by gradient descent on an image similarity through a differentiable renderer. The SOD is the hardest parameter to recover, with a reported residual error of $\pm15\,\mathrm{mm}$~\cite{gopalakrishnan2026rapid}. In our data, the estimated SOD placed parts of the lungs behind the detector for $11.7\%$ of the PA and $54.8\%$ of the lateral X-rays. We therefore adapt xvr in two ways. First, rotations are defined around the lung center instead of the CT center. Second, the SOD $y$ is restricted to $\pm 20\,\mathrm{mm}$ around the value $y_0$ expected from the acquisition geometry (SDD minus detector-to-skin gap, chest-wall thickness and half the lung extent) by optimizing a free parameter $b$ with $y = y_0 + 20\,\mathrm{mm} \cdot \tanh(b / 20\,\mathrm{mm})$. 

\paragraph{Pose head architecture.}
We use a weight-shared ResNet-18~\cite{he2016deep} to encode $x_{\text{PA}}$ and $x_{\text{lat}}$ into feature vectors $f_{\text{PA}}, f_{\text{lat}} \in \mathbb{R}^{512}$. Since pose is largely determined by the anatomy's position and apparent size on the detector, we explicitly provide these quantities in metric units. This information cannot be reliably inferred from image features alone, as pixel spacing varies across acquisitions. Using per-view organ segmentations from FleXray~\cite{butoi2026flexray}, we compute for the lung, ribs, heart, and vertebrae seven measures relative to the detector center: centroid row and column, top, bottom, left and right edge, and the area ($2{\times}28$ values). The X-ray features and mask statistics are then concatenated into a $1080$-dimensional vector and fed into an \emph{MLP} (see~\cref{fig:architecture}), which outputs the 6~DoF per view and a scale factor $k$.

\paragraph{Scale factor.}
Lung size varies strongly between patients, and deforming a fixed-size template would require large vertex displacements just to match it. The pose head therefore also predicts an isotropic scale factor $k = (V_{\text{patient}} / V_{\text{template}})^{1/3}$ that is applied to the template beforehand, so that the deformation only has to model differences in shape.

\subsection{Differentiable back-projection}
\label{sec:diff_back}
Both X-rays are lifted into a volume in the canonical frame using the cameras predicted by the pose head. Each view is modeled as a point source with a planar detector. In camera coordinates, with the origin at the source, the detector of the resized $512\times512$ X-ray is given by the vector $o_c$ to the center of pixel $(0,0)$ and the pixel steps $e_u = r\delta\,\hat{u}$ and $e_v = r\delta\,\hat{v}$, where $\delta$ is the IPS and $r$ the resize factor. The standard PA or lateral camera has rotation $R_0$ and places the lung center $c$ at position $q_0$ in camera coordinates. The pose head predicts a rotation vector $\omega$ and a translation $\Delta q$ as the deviation from this standard camera, which gives the source and detector in the canonical frame as
\begin{equation}
\begin{aligned}
R &= R_0 \exp([\omega]_\times), & s &= c - R\,(q_0 + \Delta q),\\
O &= s + R\,o_c, & E_u &= R\,e_u, \quad E_v = R\,e_v,
\end{aligned}
\end{equation}
with source position $s$ and center $O$ of pixel $(0,0)$. The volume is a grid of $128^3$ voxels covering $352\times288\times352$\,mm (left-right $\times$ anterior-posterior $\times$ superior-inferior) around $c$. For every voxel position $w$, the ray from $s$ through $w$ is intersected with the detector plane with unit normal $n = E_u \times E_v / \lVert E_u \times E_v \rVert$:
\begin{equation}
\begin{aligned}
Q &= s + \lambda (w - s), & \lambda &= \frac{(O - s)\cdot n}{(w - s)\cdot n},\\
u &= \frac{(Q - O)\cdot E_u}{\lVert E_u \rVert^2}, & v &= \frac{(Q - O)\cdot E_v}{\lVert E_v \rVert^2}.
\end{aligned}
\end{equation}
Here $Q$ is the intersection point, $\lVert \cdot \rVert$ the Euclidean norm, and $(u, v)$ its continuous column and row coordinates, at which the X-ray is sampled bilinearly. Rays that miss the image contribute zero. The two views are summed, and the volume is min-max normalized to $[0, 1]$. 

\subsection{Template deformation model}
\label{sec:template_deformation}

The architecture of the template-deformation model in SLUSH is inspired by works such as Pixel2Mesh~\cite{wang2018pixel2mesh} and Vox2Cortex~\cite{bongratz2022vox2cortex}. However, in contrast to these works that use real RGB and medical images, respectively, we use the pose-aware back-projection as input to the 3D U-Net~\cite{Isensee2020nnunet}. These features are then mapped to the vertex coordinates of the input template in canonical space, where a deep GNN predicts vertex-wise deformations. The U-Net is further trained to predict a voxel-wise segmentation map of the back-projection, which serves as an auxiliary task to learn expressive image features. SLUSH uses a dedicated lung template, obtained from a Taubin-smoothed average of the ground-truth lung meshes from 50 training subjects. It is scaled isotropically by the predicted $\hat k$ before the first deformation step. Gradients of the mesh and voxel losses reach the pose head through the differentiable back-projection and $\hat k$ through the template scaling, so pose, scale, and shape are trained jointly. Architecture details are given in Supp.~\cref{supp:implementation_details}.


\noindent The model is trained end-to-end with
\begin{equation}
\label{eq:loss}
\begin{split}
\mathcal{L} ={}& \mathcal{L}_{\text{CE}}
+ \sum_{s=1}^{4}\sum_{c\in\{\text{L},\text{R}\}}\bigl(4\,\mathcal{L}_{\text{Ch}}
+ 0.01\,\mathcal{L}_{\text{n}} + 0.1\,\mathcal{L}_{\text{lap}} \\
&+ 0.1\,\mathcal{L}_{\text{nc}} + 5\,\mathcal{L}_{\text{edge}}
+ 0.1\,\mathcal{L}_{\text{vol}}\bigr)
+ 5\,\mathcal{L}_{\text{pose}} + 2\,\mathcal{L}_{k}.
\end{split}
\end{equation}

We supervise the model with three groups of losses. For the \textbf{shape}, $\mathcal{L}_{\text{CE}}$ is the cross-entropy between the voxel decoder output and the voxelized reference labels, $\mathcal{L}_{\text{Ch}}$ the curvature-weighted Chamfer distance~\cite{bongratz2022vox2cortex} between the predicted vertices and points sampled from the reference surface, with curvature weights capped at 3, and $\mathcal{L}_{\text{n}}$ the cosine distance between the normals of matched points. For \textbf{regularization}, $\mathcal{L}_{\text{lap}}$ penalizes the Laplacian of the vertex displacements, $\mathcal{L}_{\text{nc}}$ differences between the normals of adjacent faces, and $\mathcal{L}_{\text{edge}}$ the mean squared edge length. For \textbf{pose and size}, $\mathcal{L}_{\text{pose}}$ is a weighted smooth-$\ell_1$ loss between the predicted and reference poses of both views on standardized values, $\mathcal{L}_{k}=\lvert\hat k-k\rvert$ supervises the template scale, and $\mathcal{L}_{\text{vol}}=\min\bigl(\lvert\hat V-V_{\text{GT}}\rvert/V_{\text{GT}},\,1\bigr)$ the relative volume, capped at 1 because early, self-intersecting meshes have meaningless volumes.

\subsection{Datasets and LIDC-RealPairs benchmark}
\label{sec:data}

\paragraph{In-house data.}
We train on an anonymized dataset retrospectively collected from a university hospital, with 775 training, 244 validation, and 200 held-out test pairs, split by patient. We include patients with bi-planar X-rays acquired within two days of a chest CT that fully contains the lungs. Lungs are segmented from the CT with TotalSegmentator~\cite{wasserthal2023totalsegmentator} and converted to surface meshes. CTs are rigidly aligned to a chest template, which defines the canonical frame (Supp.~\cref{supp:preprocessing}).

\paragraph{LIDC-RealPairs.}
\label{sec:benchmark}
LIDC-IDRI~\cite{armato_iii_lung_2011} is a public, de-identified dataset and was collected for lung nodule detection and is widely used for CT reconstruction from X-rays, typically through DRRs rendered from its CT volumes~\cite{ying2019x2ct}. As a by-product, it contains clinical chest X-rays for a subset of patients, which have occasionally been used for CT reconstruction~\cite{rath2026axon, cahan2025x} but were never curated for 3D reconstruction. They vary in modality and view coverage, include post-processed dual-energy images, are partly mis-oriented, and carry inconsistent geometric metadata. Of the 290 patients with X-rays, we keep those with a PA and a lateral view, a recorded SDD, and lungs fully contained in the CT, select the image with the most complete lung coverage per view, and correct its orientation and pixel spacing (Supp.~\cref{supp:benchmark}). The resulting 179 bi-planar pairs are upright DX acquisitions from a single vendor, paired with separately acquired supine CTs. The time interval between both is not recorded, but for 97 of the 98 pairs with known patient age, the same age in both studies bounds it to less than one year. We release the X-rays with their acquisition geometry, ground-truth lung meshes, and the evaluation protocol described below. LIDC-RealPairs is processed identically to the in-house data and serves exclusively as an external test set.
\section{Experiments}
\label{sec:experiments}

\subsection{Experimental setting}
The ground truth is defined in the canonical frame, which is tied to the lung center and not to the X-rays. Only our method predicts how the X-rays relate to this frame, so an offset between a prediction and the ground truth is not comparable across methods. We therefore align each prediction to the ground truth by rigid point-to-point ICP and report the average symmetric surface distance (ASSD), the 95th percentile Hausdorff distance (HD95), and the Dice score. ICP estimates rotation and translation but no scaling, since lung size is an anatomical property and scaling would conceal volume errors. To evaluate all methods on the same representation, voxel-based and implicit outputs are converted to surface meshes with marching cubes~\cite{lorensen1998marching} (Supp.~\cref{supp:preprocessing}).

For each metric, we test whether our method outperforms the strongest baseline using a one-sided paired Wilcoxon signed-rank test with Holm correction across metrics and datasets.

The canonical volume of $128^3$ voxels covers $352\times288\times352$\,mm (left-right $\times$ anterior-posterior $\times$ superior-inferior), i.e.\ a voxel size of $2.75\times2.25\times2.75$\,mm. All remaining parameters and the run time are given in Supp.~\cref{supp:implementation_details}.

\subsection{Comparison with baselines}
\label{sec:comparison_related_work}
\begin{table*}
  \caption{Shape and volume error for whole lung. Shape metrics after rigid alignment (rotation + translation); mean $\pm$ SD over cases. $^{\dagger}$~significantly better than the strongest baseline of the column (one-sided paired Wilcoxon signed-rank test, Holm-corrected, $p<0.05$).}
  \label{tab:whole_lung_shape_volume}
  \centering
  \setlength{\tabcolsep}{3pt}
  \resizebox{\textwidth}{!}{%
  \begin{tabular}{@{}lcccccccccccc@{}}
    \toprule
     & \multicolumn{6}{c}{In-house test (n=200)} & \multicolumn{6}{c}{LIDC-RealPairs (n=179)} \\
    \cmidrule(lr){2-7} \cmidrule(lr){8-13}
    Method & ASSD [mm] $\downarrow$ & HD95 [mm] $\downarrow$ & MeshDice $\uparrow$ & MAE [L] $\downarrow$ & MAPE [\%] $\downarrow$ & Pearson r $\uparrow$ & ASSD [mm] $\downarrow$ & HD95 [mm] $\downarrow$ & MeshDice $\uparrow$ & MAE [L] $\downarrow$ & MAPE [\%] $\downarrow$ & Pearson r $\uparrow$ \\
    \midrule
    \multicolumn{13}{c}{\textit{3D reconstruction models}} \\
    Mean Template & 8.27\,{\scriptsize$\pm$2.36} & 24.18\,{\scriptsize$\pm$7.66} & 0.787\,{\scriptsize$\pm$0.067} & 1.104\,{\scriptsize$\pm$0.721} & 27.80\,{\scriptsize$\pm$25.00} & N/A & 8.20\,{\scriptsize$\pm$2.57} & 23.59\,{\scriptsize$\pm$9.32} & 0.789\,{\scriptsize$\pm$0.068} & 1.017\,{\scriptsize$\pm$0.822} & 24.11\,{\scriptsize$\pm$21.83} & N/A \\
    X2CT-GAN~\cite{ying2019x2ct} & 10.41\,{\scriptsize$\pm$2.97} & 45.44\,{\scriptsize$\pm$21.56} & 0.658\,{\scriptsize$\pm$0.091} & 1.574\,{\scriptsize$\pm$0.917} & 33.56\,{\scriptsize$\pm$15.17} & 0.631 & 10.73\,{\scriptsize$\pm$3.07} & 46.66\,{\scriptsize$\pm$21.23} & 0.653\,{\scriptsize$\pm$0.088} & 1.709\,{\scriptsize$\pm$1.038} & 35.40\,{\scriptsize$\pm$14.97} & 0.540 \\
    AXON~\cite{rath2026axon} & 7.92\,{\scriptsize$\pm$1.93} & 24.01\,{\scriptsize$\pm$8.92} & 0.788\,{\scriptsize$\pm$0.059} & 0.851\,{\scriptsize$\pm$0.582} & 22.05\,{\scriptsize$\pm$20.42} & 0.643 & 9.72\,{\scriptsize$\pm$3.08} & 25.07\,{\scriptsize$\pm$7.48} & 0.750\,{\scriptsize$\pm$0.086} & 1.443\,{\scriptsize$\pm$0.962} & 38.38\,{\scriptsize$\pm$34.76} & 0.547 \\
    SSM Silhouette-Fit & 7.66\,{\scriptsize$\pm$2.17} & 19.62\,{\scriptsize$\pm$5.07} & 0.802\,{\scriptsize$\pm$0.065} & 0.833\,{\scriptsize$\pm$0.624} & 23.03\,{\scriptsize$\pm$22.43} & 0.813 & 7.80\,{\scriptsize$\pm$1.90} & 19.41\,{\scriptsize$\pm$4.69} & 0.800\,{\scriptsize$\pm$0.057} & 0.860\,{\scriptsize$\pm$0.556} & 22.54\,{\scriptsize$\pm$19.53} & \textit{0.850} \\
    DeepSSM~\cite{bhalodia2018deepssm} & 6.75\,{\scriptsize$\pm$1.93} & 18.62\,{\scriptsize$\pm$5.12} & 0.822\,{\scriptsize$\pm$0.060} & 0.638\,{\scriptsize$\pm$0.534} & 17.02\,{\scriptsize$\pm$18.73} & 0.804 & 6.69\,{\scriptsize$\pm$1.84} & 18.56\,{\scriptsize$\pm$5.68} & 0.824\,{\scriptsize$\pm$0.055} & 0.617\,{\scriptsize$\pm$0.468} & 14.88\,{\scriptsize$\pm$13.56} & 0.817 \\
    X2B~\cite{guven2026x2br} & 6.21\,{\scriptsize$\pm$2.04} & \textit{17.40}\,{\scriptsize$\pm$5.39} & \textit{0.840}\,{\scriptsize$\pm$0.061} & 0.786\,{\scriptsize$\pm$0.609} & 21.37\,{\scriptsize$\pm$20.78} & 0.838 & 6.55\,{\scriptsize$\pm$2.28} & \textit{16.52}\,{\scriptsize$\pm$5.73} & \textit{0.832}\,{\scriptsize$\pm$0.065} & 1.091\,{\scriptsize$\pm$0.640} & 28.07\,{\scriptsize$\pm$20.88} & 0.849 \\
    IGCN+~\cite{nakao2022igcn} & 7.11\,{\scriptsize$\pm$1.66} & 22.45\,{\scriptsize$\pm$6.51} & 0.816\,{\scriptsize$\pm$0.051} & 0.765\,{\scriptsize$\pm$0.541} & 19.08\,{\scriptsize$\pm$17.38} & 0.729 & 7.03\,{\scriptsize$\pm$1.83} & 21.85\,{\scriptsize$\pm$6.36} & 0.820\,{\scriptsize$\pm$0.052} & 0.774\,{\scriptsize$\pm$0.583} & 17.77\,{\scriptsize$\pm$14.57} & 0.721 \\
    X2Shape~\cite{PAN2026104085} & 6.90\,{\scriptsize$\pm$2.01} & 19.65\,{\scriptsize$\pm$6.36} & 0.808\,{\scriptsize$\pm$0.061} & 0.722\,{\scriptsize$\pm$0.580} & 17.43\,{\scriptsize$\pm$15.31} & 0.784 & 7.13\,{\scriptsize$\pm$2.10} & 19.30\,{\scriptsize$\pm$6.62} & 0.805\,{\scriptsize$\pm$0.062} & 0.750\,{\scriptsize$\pm$0.648} & 18.11\,{\scriptsize$\pm$16.36} & 0.788 \\
    Swin-X2S~\cite{liu2025swin} & 6.81\,{\scriptsize$\pm$1.78} & 21.28\,{\scriptsize$\pm$5.89} & 0.828\,{\scriptsize$\pm$0.056} & 0.717\,{\scriptsize$\pm$0.514} & 18.96\,{\scriptsize$\pm$18.80} & 0.777 & 6.85\,{\scriptsize$\pm$1.89} & 20.99\,{\scriptsize$\pm$5.76} & 0.829\,{\scriptsize$\pm$0.057} & 0.734\,{\scriptsize$\pm$0.578} & 18.56\,{\scriptsize$\pm$19.05} & 0.737 \\
    pixelNeRF~\cite{yu2021pixelnerf} & 6.61\,{\scriptsize$\pm$1.93} & 19.71\,{\scriptsize$\pm$4.74} & 0.832\,{\scriptsize$\pm$0.059} & 0.834\,{\scriptsize$\pm$0.645} & 22.95\,{\scriptsize$\pm$22.62} & 0.816 & 6.90\,{\scriptsize$\pm$2.02} & 19.49\,{\scriptsize$\pm$4.70} & 0.828\,{\scriptsize$\pm$0.061} & 1.030\,{\scriptsize$\pm$0.680} & 27.29\,{\scriptsize$\pm$23.24} & 0.816 \\
    Pixel2Mesh++~\cite{wen2019pixel2mesh++} & \textit{6.14}\,{\scriptsize$\pm$1.56} & 18.18\,{\scriptsize$\pm$5.61} & 0.829\,{\scriptsize$\pm$0.050} & 0.601\,{\scriptsize$\pm$0.455} & 14.40\,{\scriptsize$\pm$12.66} & 0.828 & \textit{6.37}\,{\scriptsize$\pm$1.92} & 17.66\,{\scriptsize$\pm$6.04} & 0.825\,{\scriptsize$\pm$0.058} & 0.679\,{\scriptsize$\pm$0.538} & 16.80\,{\scriptsize$\pm$17.65} & 0.757 \\
    SPAR3D~\cite{huang2025spar3d} & 7.05\,{\scriptsize$\pm$2.40} & 18.26\,{\scriptsize$\pm$5.99} & 0.807\,{\scriptsize$\pm$0.076} & 0.730\,{\scriptsize$\pm$0.581} & 19.66\,{\scriptsize$\pm$20.05} & 0.760 & 6.96\,{\scriptsize$\pm$2.10} & 17.46\,{\scriptsize$\pm$5.08} & 0.813\,{\scriptsize$\pm$0.065} & 0.695\,{\scriptsize$\pm$0.492} & 18.04\,{\scriptsize$\pm$16.63} & 0.821 \\
    \textbf{SLUSH (ours)} & \textbf{4.87}$^{\dagger}$\,{\scriptsize$\pm$1.60} & \textbf{14.62}$^{\dagger}$\,{\scriptsize$\pm$5.70} & \textbf{0.869}$^{\dagger}$\,{\scriptsize$\pm$0.049} & \textit{0.554}\,{\scriptsize$\pm$0.435} & 14.04\,{\scriptsize$\pm$14.48} & \textit{0.850} & \textbf{4.42}$^{\dagger}$\,{\scriptsize$\pm$1.30} & \textbf{12.58}$^{\dagger}$\,{\scriptsize$\pm$3.60} & \textbf{0.882}$^{\dagger}$\,{\scriptsize$\pm$0.041} & \textbf{0.474}$^{\dagger}$\,{\scriptsize$\pm$0.385} & \textbf{11.66}$^{\dagger}$\,{\scriptsize$\pm$11.73} & \textbf{0.891} \\
    \midrule
    \multicolumn{13}{c}{\textit{Volume prediction models (no shape output)}} \\
    Barnhard \etal~\cite{barnhard1960roentgenographic} & N/A & N/A & N/A & 0.648\,{\scriptsize$\pm$0.471} & 16.10\,{\scriptsize$\pm$14.90} & 0.795 & N/A & N/A & N/A & \textit{0.613}\,{\scriptsize$\pm$0.506} & 14.40\,{\scriptsize$\pm$13.25} & 0.793 \\
    Dorosti \etal~\cite{dorosti2025estimating} & N/A & N/A & N/A & \textbf{0.522}\,{\scriptsize$\pm$0.432} & \textbf{13.06}\,{\scriptsize$\pm$13.29} & \textbf{0.858} & N/A & N/A & N/A & 0.637\,{\scriptsize$\pm$0.450} & \textit{14.22}\,{\scriptsize$\pm$10.54} & 0.834 \\
    Ghimire and Subedi~\cite{ghimire2024estimating} & N/A & N/A & N/A & 0.595\,{\scriptsize$\pm$0.500} & 15.23\,{\scriptsize$\pm$15.72} & 0.820 & N/A & N/A & N/A & 0.719\,{\scriptsize$\pm$0.548} & 17.61\,{\scriptsize$\pm$16.91} & 0.745 \\
    Sogancioglu \etal~\cite{sogancioglu2022automated} & N/A & N/A & N/A & 0.565\,{\scriptsize$\pm$0.437} & \textit{13.76}\,{\scriptsize$\pm$13.19} & 0.842 & N/A & N/A & N/A & 0.709\,{\scriptsize$\pm$0.557} & 15.75\,{\scriptsize$\pm$12.71} & 0.756 \\
    \bottomrule
  \end{tabular}}
\end{table*}


We compare against representative methods from various reconstruction paradigms. All learned baselines are trained on the same in-house split, and single-view methods are adapted to bi-planar X-rays (Supp.~\cref{supp:baselines}). As shown in \cref{tab:whole_lung_shape_volume}, SLUSH is significantly better on all shape metrics on both datasets.

The CT synthesis models X2CT-GAN~\cite{ying2019x2ct} and AXON~\cite{rath2026axon}, followed by lung segmentation of the synthesized CT, perform worst, which is plausible as they reconstruct the full attenuation volume rather than a specific organ shape. An SSM fitted to lung silhouettes predicted by FleXray \cite{butoi2026flexray} shows what silhouettes alone provide. DeepSSM \cite{bhalodia2018deepssm}, the mesh-based IGCN+ \cite{nakao2022igcn}, and the voxel-based X2Shape \cite{PAN2026104085} and Swin-X2S \cite{liu2025swin} perform similarly, and the occupancy-based X2B \cite{guven2026x2br} is the strongest of the X-ray-specific baselines. The retrained computer vision models pixelNeRF \cite{yu2021pixelnerf}, Pixel2Mesh++ \cite{wen2019pixel2mesh++}, and SPAR3D \cite{huang2025spar3d} are competitive with these. Pixel2Mesh++, like our method a template deformation model with calibrated projection, achieves the lowest ASSD among all baselines. None of them estimates the pose, however, and SLUSH lowers the ASSD of the strongest baseline by 20.7\%/30.6\% and improves the MeshDice by 3.5\%/6.0\% on the in-house test set and LIDC-RealPairs. 

To understand the larger margin on LIDC-RealPairs, we replace the predicted cameras at test time by registration-based reference cameras, which require the CT (Supp.~\cref{supp:oracle_analysis}). With these, both datasets are equally difficult. On LIDC-RealPairs, however, the predicted cameras already yield reconstructions as accurate as the reference cameras, while a gap remains on the in-house data.

\paragraph{Downstream volumetry.}
As a clinically relevant downstream task, we estimate total lung volume from the reconstructed meshes and compare it with the CT-derived reference. Following prior work~\cite{sogancioglu2022automated}, we report the mean absolute error (MAE), the mean absolute percentage error (MAPE), and the Pearson correlation coefficient. In addition to the reconstruction baselines, we compare against dedicated volume estimation methods: the classical ellipse method \cite{barnhard1960roentgenographic} and the regression approaches of Sogancioglu~\etal~\cite{sogancioglu2022automated}, Ghimire and Subedi~\cite{ghimire2024estimating}, and Dorosti~\etal~\cite{dorosti2025estimating}. On the in-house test set, SLUSH performs best among the reconstruction models but does not outperform Dorosti~\etal. On LIDC-RealPairs, it is significantly better than all methods, lowering the MAE by 22.7\% against the ellipse method and the MAPE by 18\% against Dorosti~\etal.


\begin{figure*}
    \centering
    \includegraphics[width=\textwidth]{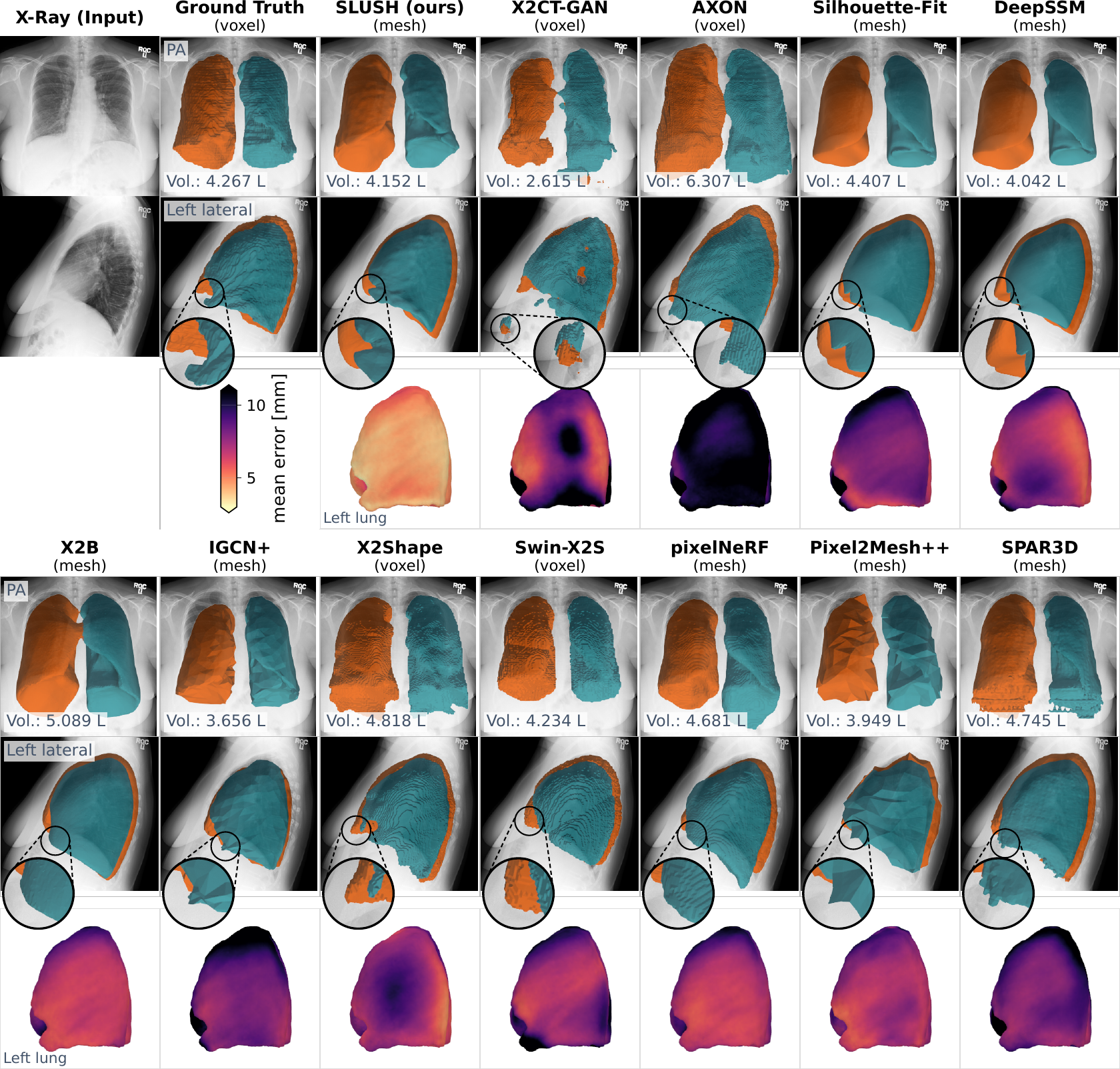}
    \caption{Qualitative comparison based on an exemplary pair of bi-planar X-rays from the LIDC-RealPairs benchmark (LIDC-IDRI-0265). Predicted right (orange) and left (teal) lungs are overlaid on the input X-rays in the frame in which each method predicts them. The bottom row shows the surface error of the left lung over the whole LIDC-RealPairs benchmark ($n=179$).}
\label{fig:qualitative}
\end{figure*}

\cref{fig:qualitative} shows an example from LIDC-RealPairs, with zoom-ins on the posterior costophrenic recess and the average surface error of the left lung over the LIDC-RealPairs benchmark. For X2CT-GAN and AXON, lungs are segmented from the synthesized CT with TotalSegmentator~\cite{wasserthal2023totalsegmentator}. X2CT-GAN yields fragmented lungs with holes and a far too low volume (2.6\,L vs.\ 4.3\,L), while AXON produces coherent lungs but overestimates the volume by about 2\,L. The SSM-based Silhouette-Fit and DeepSSM give smooth lungs of roughly correct volume but cannot represent the thin recess. X2B overestimates the volume by about 0.8\,L, and IGCN+ and Pixel2Mesh++ show coarse, creased surfaces. The voxel-based X2Shape and Swin-X2S exhibit the staircase pattern inherent to voxel outputs, pixelNeRF overestimates the volume by about 0.4\,L, and SPAR3D leaves scattered fragments at the diaphragm. SLUSH reconstructs smooth lungs that follow the lung fields in both views, including the recess, and its error map is low over the entire surface, whereas all baselines show large errors at the lung base.

The supplement provides further examples, including a failure case with inconsistent PA and lateral views (Supp.~\cref{supp:additional_results}), and multi-organ results (Supp.~\cref{supp:multiple_organ}). Adding the reconstructed meshes to an image-based classifier raises the macro AUROC on nine shape-related findings from 0.813 to 0.839, with gains on all nine (Supp.~\cref{supp:diagnostic_information}).

\subsection{Ablation study}
\label{sec:ablation}

\begin{table}
  \caption{Additive ablation: starting from naive bi-planar input, each row adds one component to the row above. Mean $\pm$ SD over the 244 validation cases.}
  \label{tab:ablation_ladder}
  \centering
  \setlength{\tabcolsep}{3pt}
  \resizebox{\columnwidth}{!}{%
  \begin{tabular}{@{}lcccccc@{}}
    \toprule
    Step & Rot $\downarrow$ & Cam $\downarrow$ & ASSD $\downarrow$ & HD95 $\downarrow$ & Dice $\uparrow$ & Vol.\ err.\ $\downarrow$ \\
     & [$^\circ$] & [mm] & [mm] & [mm] & & [\%] \\
    \midrule
    \makecell[lc]{bi-planar,\\naive input} & 10.37$^{*}$\,{\scriptsize$\pm$4.53} & 157.60$^{*}$\,{\scriptsize$\pm$14.82} & 6.39\,{\scriptsize$\pm$1.99} & 19.12\,{\scriptsize$\pm$6.45} & 0.828\,{\scriptsize$\pm$0.057} & 17.67\,{\scriptsize$\pm$20.52} \\
    \midrule
    \makecell[lc]{+ back-projection\\with pose head} & 8.91\,{\scriptsize$\pm$4.08} & 26.03\,{\scriptsize$\pm$9.57} & 5.97\,{\scriptsize$\pm$1.87} & 18.70\,{\scriptsize$\pm$7.10} & 0.840\,{\scriptsize$\pm$0.052} & 15.99\,{\scriptsize$\pm$17.09} \\
    \midrule
    + pose loss & 8.48\,{\scriptsize$\pm$4.05} & 26.85\,{\scriptsize$\pm$9.75} & 5.86\,{\scriptsize$\pm$1.86} & 18.79\,{\scriptsize$\pm$7.35} & 0.842\,{\scriptsize$\pm$0.052} & 16.04\,{\scriptsize$\pm$17.12} \\
    \midrule
    + $k$ head & 8.45\,{\scriptsize$\pm$3.95} & 24.71\,{\scriptsize$\pm$9.27} & 5.80\,{\scriptsize$\pm$1.83} & 17.91\,{\scriptsize$\pm$6.89} & 0.843\,{\scriptsize$\pm$0.053} & 16.14\,{\scriptsize$\pm$17.20} \\
    \midrule
    \makecell[lc]{+ warm-started\\pose head} & \textbf{7.13}\,{\scriptsize$\pm$3.40} & 19.93\,{\scriptsize$\pm$7.78} & 5.08\,{\scriptsize$\pm$1.66} & 15.83\,{\scriptsize$\pm$6.24} & 0.862\,{\scriptsize$\pm$0.046} & 14.26\,{\scriptsize$\pm$14.46} \\
    \midrule
    + mask features & 7.25\,{\scriptsize$\pm$3.44} & 19.48\,{\scriptsize$\pm$8.01} & 5.07\,{\scriptsize$\pm$1.69} & 15.71\,{\scriptsize$\pm$6.53} & 0.862\,{\scriptsize$\pm$0.046} & 14.23\,{\scriptsize$\pm$14.54} \\
    \midrule
    \makecell[lc]{+ intensity\\handling} & 7.15\,{\scriptsize$\pm$3.45} & \textbf{19.41}\,{\scriptsize$\pm$8.24} & 5.00\,{\scriptsize$\pm$1.69} & 15.29\,{\scriptsize$\pm$6.30} & 0.864\,{\scriptsize$\pm$0.047} & 14.06\,{\scriptsize$\pm$14.95} \\
    \midrule
    \makecell[lc]{+ volume loss\\(final)} & 7.21\,{\scriptsize$\pm$3.43} & 19.44\,{\scriptsize$\pm$8.00} & \textbf{4.96}\,{\scriptsize$\pm$1.70} & \textbf{14.94}\,{\scriptsize$\pm$6.24} & \textbf{0.865}\,{\scriptsize$\pm$0.047} & \textbf{13.92}\,{\scriptsize$\pm$14.92} \\
    \bottomrule
  \end{tabular}}
\end{table}

\cref{tab:ablation_ladder} adds the components of SLUSH one at a time. The naive baseline lifts both X-rays by repeating them along their view direction. This assumes that the X-rays were acquired from the canonical views, whereas the xvr reference cameras deviate from these by about $10^\circ$ and 16\,cm on average. Back-projecting through a pose head instead improves the shape. Notably, the mesh losses alone, without any pose supervision, reduce the camera error from 157.6 to 26.0\,mm. Adding the xvr references changes little beyond this, so the pose is learned from the reconstruction objective. Predicting the template scale $k$ separates patient size from depth and mainly improves the camera position. Its error is only moderately correlated with the depth error ($r = 0.38$) and far too small to trade off against it. Warm-starting the pose head for 100 epochs is the largest single step (ASSD $-0.72$\,mm, rotation $-1.3^\circ$). Mask features, intensity handling, and the volume loss each add small but consistent gains, and the final model is best on all reconstruction metrics.

A further analysis (Supp.~\cref{supp:oracle_analysis}) examines what the model could achieve with ground-truth pose and scale.

\subsection{2D-to-3D nodule localization}
\label{sec:nodule_localization}

\begin{table}
  \caption{2D-to-3D nodule localization on 26 nodules from LIDC-RealPairs.
  Baselines use template cameras, ours its predicted cameras. All metrics are
  computed after rigid ICP onto the reference lungs. 
  Bottom: ground-truth (GT) lungs with template cameras.
  Hits: estimates that overlap the CT nodule (Dice $>0$). Mean $\pm$ SD over nodules.}
  \label{tab:nodule_backprojection_template_icp_hull}
  \centering
  \setlength{\tabcolsep}{3pt}
  \resizebox{\columnwidth}{!}{%
  \begin{tabular}{@{}lcccccc@{}}
    \toprule
    Method & Hits $\uparrow$ & Dice $\uparrow$ & IoU $\uparrow$ & \makecell{Med. Dist.\ $\downarrow$ \\ {[mm]}} & \makecell{Vol.\ err.\ $\downarrow$ \\ {[mL]}} & Pearson $r$ $\uparrow$ \\
      \midrule
    X2CT-GAN~\cite{ying2019x2ct} & 4/26 & 0.034\,{\scriptsize$\pm$0.118} & 0.02\,{\scriptsize$\pm$0.08} & 30.13 & 5.00\,{\scriptsize$\pm$2.90} & 0.46 \\
    AXON~\cite{rath2026axon} & 7/26 & 0.017\,{\scriptsize$\pm$0.045} & 0.01\,{\scriptsize$\pm$0.02} & 29.93 & 4.82\,{\scriptsize$\pm$2.61} & 0.65 \\
    SSM Silhouette-Fit & 6/26 & 0.035\,{\scriptsize$\pm$0.090} & 0.02\,{\scriptsize$\pm$0.05} & 30.40 & 4.73\,{\scriptsize$\pm$2.63} & 0.71 \\
    DeepSSM~\cite{bhalodia2018deepssm} & 5/26 & 0.028\,{\scriptsize$\pm$0.077} & 0.02\,{\scriptsize$\pm$0.04} & 27.29 & 4.70\,{\scriptsize$\pm$2.51} & 0.68 \\
    X2B~\cite{guven2026x2br} & 4/26 & 0.032\,{\scriptsize$\pm$0.086} & 0.02\,{\scriptsize$\pm$0.05} & 31.07 & 4.69\,{\scriptsize$\pm$2.73} & 0.70 \\
    IGCN+~\cite{nakao2022igcn} & 5/26 & 0.022\,{\scriptsize$\pm$0.054} & 0.01\,{\scriptsize$\pm$0.03} & 33.24 & 4.74\,{\scriptsize$\pm$2.84} & 0.62 \\
    X2Shape~\cite{PAN2026104085} & 5/26 & 0.006\,{\scriptsize$\pm$0.024} & 0.00\,{\scriptsize$\pm$0.01} & 30.75 & 4.50\,{\scriptsize$\pm$2.76} & 0.71 \\
    Swin-X2S~\cite{liu2025swin} & 5/26 & 0.005\,{\scriptsize$\pm$0.015} & 0.00\,{\scriptsize$\pm$0.01} & 31.91 & 4.56\,{\scriptsize$\pm$2.78} & 0.73 \\
    pixelNeRF~\cite{yu2021pixelnerf} & 5/26 & 0.021\,{\scriptsize$\pm$0.061} & 0.01\,{\scriptsize$\pm$0.03} & 32.02 & 4.70\,{\scriptsize$\pm$2.70} & 0.69 \\
    Pixel2Mesh++~\cite{wen2019pixel2mesh++} & 6/26 & 0.023\,{\scriptsize$\pm$0.058} & 0.01\,{\scriptsize$\pm$0.03} & 32.54 & 4.87\,{\scriptsize$\pm$2.56} & 0.66 \\
    SPAR3D~\cite{huang2025spar3d} & 8/26 & 0.013\,{\scriptsize$\pm$0.051} & 0.01\,{\scriptsize$\pm$0.03} & 30.89 & 4.70\,{\scriptsize$\pm$2.53} & 0.72 \\
    SLUSH (ours) & \textbf{21/26} & \textbf{0.197}\,{\scriptsize$\pm$0.187} & \textbf{0.12}\,{\scriptsize$\pm$0.13} & \textbf{12.18} & \textbf{2.23}\,{\scriptsize$\pm$1.86} & \textbf{0.78} \\
    \midrule
    GT lungs + template cameras & 8/26 & 0.024\,{\scriptsize$\pm$0.056} & 0.01\,{\scriptsize$\pm$0.03} & 26.50 & 4.45\,{\scriptsize$\pm$3.18} & 0.36 \\
    \bottomrule
  \end{tabular}}
\end{table}

The 3D location of a nodule determines how it can be accessed, for example by biopsy or resection. On a single radiograph, this location is ambiguous, as each view only constrains the lesion to a ray. Intersecting the rays of both views resolves it, but only if the cameras are known, which makes this task a direct test of the predicted pose. We back-project nodules annotated on both X-rays into the reconstructed lungs using the predicted cameras (\cref{fig:teaser}) and compare them with the nodules in the CT (Supp.~\cref{supp:nodule_localization}). The baselines predict no cameras and therefore use the template cameras they were trained with. As shown in \cref{tab:nodule_backprojection_template_icp_hull}, SLUSH localizes 21 of 26 nodules, compared to at most 8 for the baselines, and achieves a Dice of 0.197 versus 0.035 for the best baseline (Silhouette-Fit). The median distance between estimated and true nodule center drops to 12.2\,mm, compared with 27.3\,mm (DeepSSM), and the error in nodule volume to 2.2\,mL, compared with 4.5\,mL (X2Shape). Even the ground-truth lungs reach only 8 of 26 hits with the template cameras, so the gain stems from the predicted pose and not from the shapes. 

\section{Limitations}

The CT reference is acquired supine and at a different breathing state than the upright X-rays, so part of the remaining error reflects true anatomical differences rather than reconstruction errors. The ground-truth surfaces are meshes extracted from CT segmentations, so they approximate the true lung boundary. Pose supervision relies on registration-based reference cameras, whose depth in particular remains uncertain. While the in-house data covers X-ray systems from multiple vendors, the X-rays in LIDC-RealPairs come from one vendor and from a lung-nodule population.


\section{Conclusion}
We addressed 3D lung reconstruction from real bi-planar X-rays, where input and target are not aligned by construction as they are for DRRs. Our method, SLUSH, separates pose from shape. A pose head maps both X-rays into the canonical frame of an anatomical template, which is then deformed to the patient-specific anatomy. Trained exclusively on clinical X-ray--CT pairs, SLUSH outperforms baselines from various reconstruction paradigms on an in-house test set and on LIDC-RealPairs, our new public benchmark, reducing the ASSD by $20.7\%$ and $30.6\%$ relative to the strongest baseline. Volumes derived from the meshes are competitive with dedicated regression methods in-house and more accurate on the external benchmark.
Finally, the predicted cameras localize nodules annotated on the X-rays far more reliably than any baseline.

\section*{Acknowledgements}
This research was supported by the German Research Foundation (DFG) and the Munich Center for Machine Learning (MCML). We gratefully acknowledge the computational resources provided by the Leibniz Supercomputing Centre (www.lrz.de).
\section*{LLM Usage Statement}
\label{sec:llm_statement}
The authors take full responsibility for all submitted content, regardless of any use of Large Language Models (LLMs) or generative AI tools. To ensure complete transparency, we have disclosed the specific roles of AI tools in the development and writing of this work:
\begin{itemize}
    \item \textbf{Writing and Editing:} Anthropic Claude and Google Gemini were used to edit and refine the authors text in terms of grammar, clarity and conciseness throughout the manuscript. All scientific content, claims and interpretations originate from the authors.
    \item \textbf{Feedback:} The complete manuscript was provided to Anthropic Claude for preliminary, reviewer-style feedback. All resulting revisions were decided upon and implemented by the authors.
    \item \textbf{Related Work:} Anthropic Claude was used to help identify potentially relevant literature. All cited works were retrieved, read and verified by the authors. No reference was included without manual verification.
    \item \textbf{Code and Experiments:} Anthropic Claude was used as a coding assistant. Based on the authors specifications, it generated parts of the implementation of the authors' methods, produced plotting scripts and assisted with debugging.
    \item \textbf{Figures:} Some plots were created using LLM-assisted plotting scripts from the authors' experimental results. No figure content or data was generated by an LLM.
    \item \textbf{Other Research Phases:} No LLMs were used to develop the core research ideas, formulate hypotheses, or annotate data.
\end{itemize}
{
    
    \small
    \bibliographystyle{ieeenat_fullname}
    \bibliography{main}
}

\clearpage
\maketitlesupplementary

\setcounter{section}{0}
\renewcommand{\thesection}{\Alph{section}}

\section{LIDC-RealPairs: curation and release}
\label{supp:benchmark}

\paragraph{Case selection.}
LIDC-IDRI~\cite{armato_iii_lung_2011} contains some cases with a single X-ray view only (mainly PA), so we manually picked the cases which have both PA and lateral view. Since we obtain the lungs by segmenting the CT, we excluded the cases where the lungs are not fully covered by the CT. Cases without a set SDD field in metadata were excluded as well. \cref{tab:supp_cascade} lists the number of cases after each step.

\begin{table}[t]
\centering
\small
\caption{Filtering cascade of LIDC-RealPairs. Number of cases removed and remaining after each curation step.}
\label{tab:supp_cascade}
\begin{tabular}{llrr}
\toprule
Step & Criterion & Removed & Remaining \\
\midrule
(i)   & LIDC-IDRI~\cite{armato_iii_lung_2011} patients & -- & 290 \\
      & with $\geq 1$ X-ray (DX, CR)                   &    &     \\
(ii)  & $\geq 1$ posteroanterior and                   & 94 & 196 \\
      & $\geq 1$ lateral view                          &    &     \\
(iii) & Lungs fully contained in the CT                & 15 & 181 \\
(iv)  & Recorded SDD                                   &  2 & 179 \\
      & (excludes CR without SDD)                      &    &     \\
\bottomrule
\end{tabular}
\end{table}

\paragraph{Image selection and orientation.}
If multiple images exist per view, we visually select the one with the least lung truncation. Derived dual-energy subtraction images were excluded. Some X-rays were vertically mirrored, which we corrected via visual inspection.

\paragraph{Pixel spacing correction.}
For almost all studies, the DICOM field \texttt{ImagerPixelSpacing} contained the pixel spacing of the processed image frame instead of the detector pixel spacing. Based on the detector specification, we set it to $0.2\times0.2$\,mm.

\paragraph{Acquisition geometry.}
The SDD is taken from the DICOM header. The extrinsics produced by the adapted xvr~\cite{gopalakrishnan2026rapid} setup, described in \cref{sec:pose_head}, will be included as well. They are estimates and not ground truth, and are not part of the evaluation protocol.

\paragraph{Ground truth.}
Lungs are segmented in the CT using TotalSegmentator~\cite{wasserthal2023totalsegmentator} and converted to surface meshes as described in \cref{supp:preprocessing}. These masks and meshes serve as the primary ground truth.

\subsection{Benchmark statistics}
\begin{figure*}
    \centering
    \includegraphics[width=\textwidth]{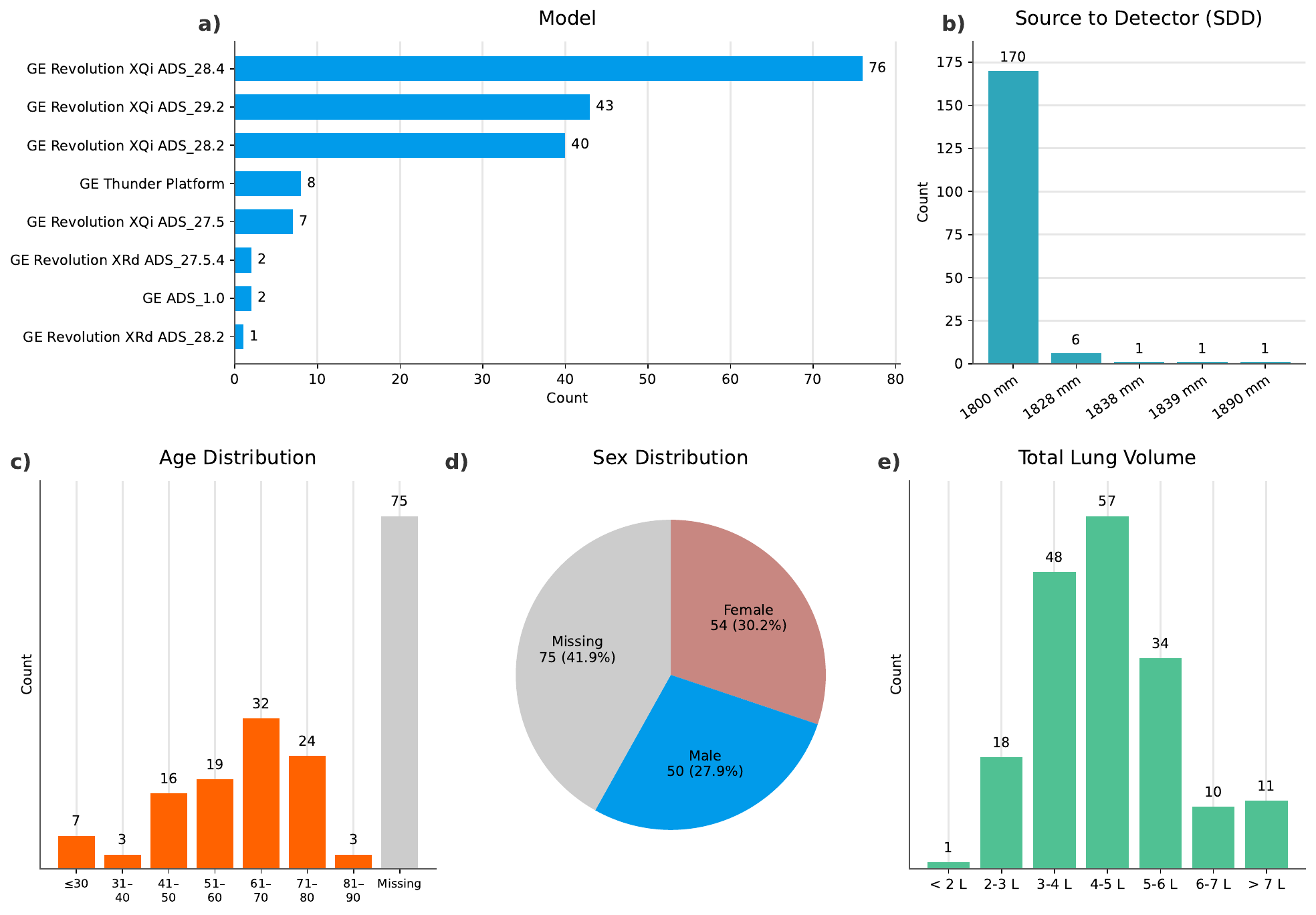}
    \caption{\textbf{Overview of the LIDC-RealPairs benchmark} (179 bi-planar X-ray--CT pairs).
    \textbf{(a)} X-ray system and software version: All X-rays are digital (DX) acquisitions from GE systems.
    \textbf{(b)} Source-to-detector distance (SDD) from the DICOM headers: 95\% of pairs use the nominal 1800\,mm.
    \textbf{(c)} Patient age and \textbf{(d)} sex from the anonymized DICOM headers of the X-ray or the CT (both are missing for the same 75 pairs and for 98 pairs, age is recorded in both studies).
    \textbf{(e)} Total lung volume (left + right) computed from the CT lung segmentation.}
    \label{fig:X_supp_benchmark_stats}
\end{figure*}

\cref{fig:X_supp_benchmark_stats} summarizes acquisition systems, geometry, demographics, and lung volumes of the benchmark.

\subsection{Release and license}
For each pair, we release the selected X-rays in DICOM format with corrected geometry metadata, the ground-truth masks and meshes, the CT series identifiers, and evaluation code. The filtered single-view X-rays are included to support single-view reconstruction. All data is derived from LIDC-IDRI \cite{ armato_iii_lung_2011, lidc_tcia_data}, which is distributed under CC BY 3.0. The modifications described above are documented, and LIDC-RealPairs is released under the same license. 

\section{Data preprocessing}
\label{supp:preprocessing}

The same preprocessing is applied to the in-house dataset and to LIDC-RealPairs.

\paragraph{CT.}
All CTs are reoriented to RAS and resampled to $1$\,mm isotropic resolution, using linear interpolation for intensities and nearest-neighbor interpolation for segmentation masks. Lungs are segmented with TotalSegmentator~\cite{wasserthal2023totalsegmentator}, and only scans with fully contained lungs are kept. To obtain a consistent canonical orientation, each CT is rigidly aligned to a chest template by maximizing mutual information on the intensity values. The world origin is centered on the lung center.

\paragraph{Mesh extraction.}
Ground-truth meshes are extracted from the lung masks in three stages:
\begin{enumerate}
  \item Mask cleanup: We keep the largest connected component and apply a morphological closing with a $3$\,mm radius to seal thin slits.
  \item Surface extraction: We compute a signed distance field in millimeters using the Euclidean distance transform, smooth it with a Gaussian of $\sigma = 1$\,mm to remove voxel staircase artifacts, and extract the zero level set with marching cubes \cite{lorensen1998marching}.
  \item Mesh processing: We keep the largest surface component and apply Taubin smoothing \cite{taubin1995signal} ($\lambda = 0.5$, $\mu = -0.53$, 12 iterations). Finally, quadric decimation \cite{garland1997surface} reduces each mesh to 16,380 faces. On a test lung, decimation changes the volume by $0.02\%$ and the surface by $0.04$\,mm ASSD. 
\end{enumerate}

\paragraph{X-rays and augmentation.}
Each view is first min-max normalized to $[0,1]$. Clinical X-rays often contain laterality markers whose intensity are to high compared to the real anatomy. With just a plain min-max normalization this compresses the anatomy into the lower part of the intensity scale. We therefore rescale each view so that the 99th percentile of its exposed pixels maps to $0.94$, and clip the result to $[0,1]$. The target value $0.94$ is the median of this percentile over the in-house training views, so typical views remain almost unchanged, while views with a marker are restored to roughly a regular contrast.

The normalized view ($H\times W$) is zero-padded at the bottom and right to a square of side $s=\max(H,W)$, which keeps pixel $(0,0)$ in place, and downsampled to $512\times512$ by area averaging over $r\times r$ pixel blocks, with $r = s/512$. To match the resampled image, the detector pixel pitch is scaled from the native spacing $\delta$ (IPS) to $r\delta$, and the detector origin is shifted by $(r-1)/2$ native pixels to the center of the first block.  

To make training robust to differences in exposure and post-processing between sites and devices, the intensities $x \in [0,1]$ of each view are independently perturbed during training as
\begin{equation}
  x' = \operatorname{clip}\!\left(g\,x^{\gamma},\, 0,\, 1\right), \qquad \gamma = 1.25^{u},
\end{equation}
where $u \sim \mathcal{U}(-1, 1)$ and $g \sim \mathcal{U}(0.9, 1.1)$ are drawn for every view and iteration. The exponent $\gamma \in [0.8, 1.25]$ alters the contrast of the mid-tones while keeping 0 and 1 fixed, and the gain $g$ changes the overall brightness by up to $\pm 10\,\%$.

The percentile normalization and this augmentation together form the \emph{intensity handling} step of the ablation in the main paper (\cref{tab:ablation_ladder}).

\section{Implementation details}
\label{supp:implementation_details}

\subsection{Architecture}

\paragraph{Pose head.}
Each view is resized to $512\times512$ pixels and encoded by a weight-shared ResNet-18~\cite{he2016deep} with a single-channel first convolution. The two 512-dimensional feature vectors are concatenated with the $2\times28$ mask statistics and passed to an MLP with layers of 512, 256 and 13 units, ReLU activations, and a dropout of 0.2 after the first layer. The 13 outputs are the six pose parameters per view and the raw scale $k_\text{raw}$, with $k = 1 + 0.4\tanh(k_\text{raw})$.

\paragraph{Voxel network.}
The back-projected volume of $128^3$ voxels is processed by a residual 3D U-Net with five resolution levels ($128^3$ to $8^3$) with 8, 16, 32, 64 and 128 encoder channels and 64, 32, 16 and 8 decoder channels, batch normalization and ReLU activations. The decoder predicts two classes (whole lung and background) for the auxiliary segmentation, with deep supervision at two intermediate decoder levels.

\paragraph{Mesh network.}
The template of each lung has 8192 vertices and 16380 faces. An initial graph block lifts the vertex coordinates and structure labels to 256 channels. The mesh is then deformed in four steps, each consisting of three residual blocks of three graph convolutions with 64 channels, batch normalization and ReLU, followed by a graph convolution that predicts the vertex displacements. U-Net features are sampled trilinearly at the vertex positions, coarse to fine: the first step uses the encoder levels at $16^3$ and $8^3$ and the decoder levels at $16^3$ and $32^3$, and each following step moves one level finer, up to encoder levels $128^3$ and $64^3$ and decoder levels $64^3$ and $128^3$ in the last step.

\subsection{Training}

\paragraph{Optimization.}
The model is implemented in PyTorch and trained with AdamW ($\beta=(0.9, 0.999)$, weight decay $10^{-4}$) and a batch size of 8. The base learning rate is $2.5\cdot10^{-4}$, and $1.25\cdot10^{-4}$ for the graph neural network, with cosine decay to $10^{-6}$ over a nominal 400 epochs and no warm-up. The model is validated every 10 epochs, and training stops 60 epochs after the lowest validation ASSD, here after 140 epochs. The checkpoint with the lowest validation ASSD is used for evaluation.

\paragraph{Warm start.}
The ResNet-18 trunk is taken from a standalone pose regressor trained for 100 epochs (best validation epoch 93) on the training cases with the pose loss only (AdamW, one-cycle schedule, maximum learning rate $3\cdot10^{-4}$, same intensity augmentation). The MLP head, which additionally takes the mask statistics and predicts $k$, is initialized from scratch. Both are then trained jointly with the reconstruction network.

\paragraph{Loss weights.}
The weights of the loss terms are given in \cref{eq:loss}. Within $\mathcal{L}_\text{pose}$, the three rotation parameters and the left-right translations are weighted by 0.42, the depth by 1.02 and the cranio-caudal translation by 3.3, for each view. The Chamfer loss compares the predicted vertices with 8192 points per lung sampled from the reference surface.

\subsection{Runtime}
Training runs on a single GPU (RTX PRO 4000 Blackwell, 24\,GB) and takes approximately 170\,s per epoch, i.e.\ about 6.6\,h for 140 epochs (9.3\,h including the intermediate validation), plus the standalone pose regressor pretraining (approximately 10 minutes). Inference takes 3\,s per case.

\section{Baseline adaptations}
\label{supp:baselines}
As far as possible, we used the original implementation for each baseline. For some baselines an adaptation was needed. All other baselines are used as published.
\begin{itemize}
  \item Sogancioglu~\etal~\cite{sogancioglu2022automated} described various architecture approaches. We used their overall best bi-planar input \emph{Dual CNN} approach. 

  \item IGCN+ \cite{nakao2022igcn} was intended for single view. For our adaptation an independent U-Net processes the lateral film. Both predicted displacement maps are sampled at the projected template vertices and concatenated.

  \item For X2B \cite{guven2026x2br} both views are fed into the same ConvNeXt encoder (shared weights). The pooled features in the end are concatenated. We used the X2B part of the model without the template mesh (X2BR). Instead of normalizing each mesh to its own size, as described in the paper, we set an anchor and a canonical frame around the center of the lung and fix the relative size to make size and position predictable.

  \item The approach proposed by Dorosti~\etal\cite{dorosti2025estimating} was adapted from one to two views by providing both views as two input channels for the U-Net and it was forced to predict two thickness maps instead of one. 
  
  \item DeepSSM \cite{bhalodia2018deepssm} was originally implemented to take a 3D volume as input. We added a back-projection of the two views from the canonical view, without registration towards the ground truth, into a 3D volume. 160 modes have been used.
  
  \item The SSM-Silhouette baseline fits a lung SSM to each case by optimization alone. It uses the same SSM as DeepSSM, but with 40 modes. The soft-rendered silhouettes of both lungs are merged into a single mask and matched to the FleXray~\cite{butoi2026flexray} lung masks of the frontal and lateral X-rays using a soft Dice loss. Pose and shape coefficients are optimized in alternation.

  \item We used the mean shape of the SSM for the mean template. Furthermore we didn't register the template and kept it at the canonical frame center. 

   \item For pixelNeRF~\cite{yu2021pixelnerf} we use the official multi-view implementation without changes to the architecture. As it is supervised by rendered views and not by 3D labels, the two X-rays serve as source views and 24 shaded renderings of the ground-truth mesh as target views, with one model per structure. A mesh is extracted by thresholding the predicted density and applying marching cubes.

  \item Pixel2Mesh++~\cite{wen2019pixel2mesh++} was re-implemented in PyTorch following the official code. Instead of an ellipsoid, each structure starts from an organ-shaped template, and all structures are predicted by one network as separate closed meshes. Each coarse block predicts vertex offsets to the previous mesh instead of absolute coordinates, as the original formulation produced folded meshes.

  \item SPAR3D~\cite{huang2025spar3d} keeps its two stages, a point diffusion model and a triplane transformer that produces the mesh. The two X-rays are encoded as two views by a frozen RAD-DINO encoder, and the network outputs one signed-distance field per organ instead of a single textured object. The per-object normalization is replaced by a fixed metric frame to preserve absolute volume. Apart from the frozen encoder, SPAR3D is trained from scratch, like all other models.
  
\end{itemize}

\section{Additional analyses}

\subsection{Gap analysis: from DRRs to real X-rays}
\label{supp:oracle_analysis}

\begin{table*}
  \caption{Gap analysis (whole lung). Each stage is derived from the SLUSH architecture and names the training and test images (DRR or real X-rays) and the geometry used. Mean ± SD over the test cases.}
 \label{tab:gap_test}
  \centering
  \setlength{\tabcolsep}{3pt}
  \resizebox{\textwidth}{!}{%
  \begin{tabular}{@{}lcccccccccccc@{}}
    \toprule
     & \multicolumn{6}{c}{In-house test (n=200)} & \multicolumn{6}{c}{LIDC-RealPairs (n=179)} \\
    \cmidrule(lr){2-7} \cmidrule(lr){8-13}
    Stage (train $\to$ test) & Vol.\ MAE [L] $\downarrow$ & Vol.\ MAPE [\%] $\downarrow$ & Vol.\ $r$ $\uparrow$ & ASSD [mm] $\downarrow$ & HD95 [mm] $\downarrow$ & Dice $\uparrow$ & Vol.\ MAE [L] $\downarrow$ & Vol.\ MAPE [\%] $\downarrow$ & Vol.\ $r$ $\uparrow$ & ASSD [mm] $\downarrow$ & HD95 [mm] $\downarrow$ & Dice $\uparrow$ \\
    \midrule
    1\ \ DRR $\to$ DRR, ideal geometry & 0.107\,{\scriptsize$\pm$0.088} & 2.69\,{\scriptsize$\pm$2.67} & 0.995 & 2.66\,{\scriptsize$\pm$0.94} & 8.75\,{\scriptsize$\pm$4.10} & 0.927\,{\scriptsize$\pm$0.031} & 0.091\,{\scriptsize$\pm$0.076} & 2.13\,{\scriptsize$\pm$1.88} & 0.996 & 2.46\,{\scriptsize$\pm$0.77} & 7.81\,{\scriptsize$\pm$2.64} & 0.934\,{\scriptsize$\pm$0.024} \\
    \midrule
    2\ \ DRR $\to$ real, canonical & 0.827\,{\scriptsize$\pm$0.643} & 18.80\,{\scriptsize$\pm$14.40} & 0.664 & 8.78\,{\scriptsize$\pm$1.84} & 29.58\,{\scriptsize$\pm$10.39} & 0.773\,{\scriptsize$\pm$0.049} & 0.757\,{\scriptsize$\pm$0.513} & 19.27\,{\scriptsize$\pm$15.96} & 0.814 & 8.38\,{\scriptsize$\pm$1.86} & 24.81\,{\scriptsize$\pm$7.94} & 0.789\,{\scriptsize$\pm$0.054} \\
    \midrule
    3\ \ real $\to$ real, naive bi-planar input & 0.660\,{\scriptsize$\pm$0.488} & 16.22\,{\scriptsize$\pm$14.59} & 0.782 & 6.15\,{\scriptsize$\pm$1.80} & 18.27\,{\scriptsize$\pm$6.20} & 0.837\,{\scriptsize$\pm$0.055} & 0.674\,{\scriptsize$\pm$0.585} & 17.07\,{\scriptsize$\pm$19.03} & 0.751 & 6.31\,{\scriptsize$\pm$2.11} & 17.79\,{\scriptsize$\pm$5.79} & 0.835\,{\scriptsize$\pm$0.060} \\
    \midrule
    4\ \ real $\to$ real, xvr pose (oracle) & 0.395\,{\scriptsize$\pm$0.334} & 10.03\,{\scriptsize$\pm$11.21} & 0.923 & 4.59\,{\scriptsize$\pm$1.44} & 14.43\,{\scriptsize$\pm$5.10} & 0.877\,{\scriptsize$\pm$0.046} & 0.426\,{\scriptsize$\pm$0.400} & 10.20\,{\scriptsize$\pm$10.09} & 0.897 & 4.62\,{\scriptsize$\pm$2.00} & 14.53\,{\scriptsize$\pm$8.94} & 0.879\,{\scriptsize$\pm$0.053} \\
    \midrule
    5\ \ real $\to$ real, xvr pose + scale (oracle) & 0.324\,{\scriptsize$\pm$0.245} & 7.89\,{\scriptsize$\pm$6.77} & 0.955 & 4.45\,{\scriptsize$\pm$1.30} & 13.92\,{\scriptsize$\pm$5.03} & 0.879\,{\scriptsize$\pm$0.045} & 0.357\,{\scriptsize$\pm$0.335} & 8.27\,{\scriptsize$\pm$7.61} & 0.931 & 4.44\,{\scriptsize$\pm$1.81} & 13.49\,{\scriptsize$\pm$6.36} & 0.883\,{\scriptsize$\pm$0.051} \\
    \midrule
    6\ \ real $\to$ real, pose head (SLUSH) & 0.554\,{\scriptsize$\pm$0.435} & 14.04\,{\scriptsize$\pm$14.48} & 0.850 & 4.87\,{\scriptsize$\pm$1.60} & 14.62\,{\scriptsize$\pm$5.70} & 0.869\,{\scriptsize$\pm$0.049} & 0.474\,{\scriptsize$\pm$0.385} & 11.66\,{\scriptsize$\pm$11.73} & 0.891 & 4.42\,{\scriptsize$\pm$1.30} & 12.58\,{\scriptsize$\pm$3.60} & 0.882\,{\scriptsize$\pm$0.041} \\
    \bottomrule
  \end{tabular}}
\end{table*}

\Cref{tab:gap_test} evaluates the SLUSH architecture in six setups, from ideal synthetic data to the real setting.

\paragraph{1 DRR $\to$ DRR, ideal geometry.}
The model is trained and tested on DRRs, as many baselines are in their original setups. DRRs are aligned with the CT and have known intrinsics and extrinsics, so no pose head is needed and the DRRs are back-projected directly into the canonical frame. They are generated with DiffDRR~\cite{gopalakrishnan2022fast}. The ASSD of 2.66\,mm and 2.46\,mm is the upper limit of what this architecture achieves from bi-planar views.

\paragraph{2 DRR $\to$ real, canonical.}
The DRR-trained model of Stage 1 is evaluated on real X-rays, assuming the canonical views. With an ASSD of 8.78\,mm and 8.38\,mm, it is weaker than SLUSH and almost all baselines (\cref{tab:whole_lung_shape_volume}), which shows the gap between DRRs and real X-rays.

\paragraph{3 Real $\to$ real, naive bi-planar input.}
The model is trained and tested on real X-rays. Instead of a back-projection, the X-rays are repeated along their view axis and concatenated.

\paragraph{4 Real $\to$ real, xvr pose (oracle).}
The model is trained on real X-rays, but the pose is not estimated. It is taken directly from xvr (for training and testing), which requires the CT. This is a reference for what our setup achieves with known extrinsics.

\paragraph{5 Real $\to$ real, xvr pose + scale (oracle).}
In addition to the oracle pose, the model receives the oracle scale factor $k$, computed from the ground-truth lung mesh.

\paragraph{6 Real $\to$ real, pose head (SLUSH).}
SLUSH as proposed in the main paper.

\paragraph{Discussion.}
On the in-house test set, SLUSH with its estimated extrinsics (4.87\,mm) comes close to the oracle pose of Stage 4 (4.59\,mm). On LIDC-RealPairs, its ASSD is even slightly lower (4.42\,mm vs.\ 4.62\,mm), a difference within the standard deviation. This explains the behavior in the main comparison (\cref{tab:whole_lung_shape_volume}). The baselines, like our canonical-view Stage 3, perform no better on LIDC-RealPairs than on the in-house test set, whereas SLUSH improves from 4.87\,mm to 4.42\,mm. With the oracle pose, both datasets are equally difficult, so the improvement comes from the pose estimation, and the extrinsics appear to be easier to estimate on LIDC-RealPairs.
The same effect is observed when the pose head is added to other architectures (\cref{tab:whole_lung_shape_volume_jointpose})

\subsection{Diagnostic information in reconstructed meshes}
\label{supp:diagnostic_information}

We test whether the reconstructed meshes carry diagnostic information, and whether they add anything to a classifier that already sees the X-rays.

\paragraph{Setup.}
We use studies from MIMIC-CXR~\cite{johnson2019mimic} with a PA and a lateral X-ray, split into $8{,}158$ studies for training and $1{,}241$ for evaluation. Labels for 26 findings are extracted from the reports with MedGemma~\cite{sellergren2025medgemma}. We compare three classifiers:
\begin{itemize}[leftmargin=*, itemsep=2pt, topsep=4pt]
  \item \textbf{Mesh based.} SLUSH reconstructs both lungs, the heart and the aorta from the two X-rays. Each mesh is compressed to a 64-dimensional shape code by a SpiralNet++ autoencoder~\cite{gong2019spiralnet++}, and the codes and relative poses of the four structures are classified by an MLP. This classifier sees no image intensities.
  \item \textbf{Image based.} Each view is processed by a DenseNet-121 pretrained on CheXpert~\cite{cohen2022torchxrayvision}. The feature maps are reduced by gated attention pooling~\cite{ilse2018attention}, concatenated, and classified by the same MLP head. The network is fine-tuned end to end.
  \item \textbf{Image + mesh based.} The outputs of the two classifiers are combined by late fusion with one weight per finding, without retraining.
\end{itemize}
We report the macro AUROC separately for nine \emph{shape-related} findings, whose radiographic sign is mainly the size, contour or position of an organ (cardiomegaly, hypoexpansion, volume loss, hyperinflated lung, flattened diaphragm, hemidiaphragm elevation, emphysema, COPD signs, mediastinal enlargement), and for the 17 other findings. Differences are tested with a patient-level bootstrap.

\paragraph{Results.}
\Cref{tab:auroc_blocks} summarizes the results, and \cref{tab:auroc_findings} lists all findings. Although it sees only shapes, the mesh based classifier reaches a macro AUROC of $0.771$ over all findings, close to the image based classifier ($0.785$). It is better on 8 of the 9 shape-related findings and worse on the others, which are mostly defined by density or texture, such as pleural effusion or consolidation.

Adding the meshes to the image based classifier raises the macro AUROC on the shape-related findings from $0.813$ to $0.839$ ($p = .002$), with gains on all nine, and leaves the other findings unchanged ($0.771$). As the meshes are reconstructed from the same X-rays, they cannot contain information that is absent from the images. They provide an explicit, compact description of organ shape, which complements an image classifier exactly where shape matters.

\begin{table}[t]
\centering
\small
\caption{Macro AUROC per group of findings.}
\label{tab:auroc_blocks}
\begin{tabular}{lccc}
\toprule
Classifier & Shape-related (9) & Others (17) & All (26) \\
\midrule
Mesh based         & .834 & .737 & .771 \\
Image based        & .813 & .771 & .785 \\
Image + mesh based & \textbf{.839} & \textbf{.771} & \textbf{.794} \\
\bottomrule
\end{tabular}
\end{table}

\begin{table*}[t]
\centering
\scriptsize
\setlength{\tabcolsep}{1.5pt}
\caption{AUROC per finding on validation + test (1,241 studies); best per column in bold. In brackets: number of positive studies. Macro: mean AUROC over the findings of the block. Shape-related: cardiomegaly, hypoexpansion, hyperinflated lung, flattened diaphragm, hemidiaphragm elevation, volume loss, emphysema, COPD signs, mediastinal enlargement.}
\label{tab:auroc_findings}
\resizebox{\textwidth}{!}{%
\begin{tabular}{lccccccccccccccccccccccccccccc}
\toprule
 & \multicolumn{10}{c}{\textbf{Shape-related (9)}} & \multicolumn{18}{c}{\textbf{Others (17)}} & \multicolumn{1}{c}{\textbf{All (26)}} \\
\cmidrule(lr){2-11} \cmidrule(lr){12-29} \cmidrule(lr){30-30}
Method & \rotatebox{90}{cardiomegaly (284)} & \rotatebox{90}{hypoexpansion (89)} & \rotatebox{90}{hyperinflated lung (79)} & \rotatebox{90}{hemidiaphragm elevation (60)} & \rotatebox{90}{emphysema (42)} & \rotatebox{90}{mediastinal enlargement (33)} & \rotatebox{90}{volume loss (31)} & \rotatebox{90}{copd signs (26)} & \rotatebox{90}{flattened diaphragm (14)} & \textbf{Macro} & \rotatebox{90}{support devices (429)} & \rotatebox{90}{pleural effusion (289)} & \rotatebox{90}{opacity unspecified (194)} & \rotatebox{90}{atelectasis unspecified (171)} & \rotatebox{90}{edema (131)} & \rotatebox{90}{fracture (105)} & \rotatebox{90}{pneumonia (82)} & \rotatebox{90}{lung lesion other (70)} & \rotatebox{90}{laminar atelectasis (67)} & \rotatebox{90}{pleural thickening (65)} & \rotatebox{90}{consolidation (63)} & \rotatebox{90}{costophrenic angle blunting (32)} & \rotatebox{90}{pneumothorax (29)} & \rotatebox{90}{loculated pleural effusion (15)} & \rotatebox{90}{mediastinal shift (15)} & \rotatebox{90}{pulmonary fibrosis (12)} & \rotatebox{90}{lobar atelectasis (11)} & \textbf{Macro} & \textbf{Macro} \\
\midrule
Mesh based classifier & .910 & \textbf{.890} & .854 & .772 & \textbf{.894} & .685 & \textbf{.811} & \textbf{.845} & .845 & .834 & .727 & .886 & .670 & .758 & .903 & .621 & .699 & \textbf{.673} & .639 & .685 & .700 & \textbf{.686} & .766 & .837 & .890 & .583 & .807 & .737 & .771 \\
Image based classifier & .902 & .817 & .868 & .762 & .847 & .666 & .803 & .819 & .831 & .813 & .739 & \textbf{.923} & .718 & .779 & .905 & .623 & .735 & .634 & \textbf{.686} & .703 & \textbf{.785} & .677 & .794 & \textbf{.917} & .896 & \textbf{.733} & \textbf{.854} & \textbf{.771} & .785 \\
Image + Mesh based classifier & \textbf{.917} & .878 & \textbf{.872} & \textbf{.782} & .890 & \textbf{.687} & \textbf{.811} & \textbf{.845} & \textbf{.869} & \textbf{.839} & \textbf{.742} & .922 & \textbf{.721} & \textbf{.780} & \textbf{.908} & \textbf{.633} & \textbf{.745} & .634 & .682 & \textbf{.705} & .784 & .681 & \textbf{.798} & \textbf{.917} & \textbf{.908} & \textbf{.733} & .807 & \textbf{.771} & \textbf{.794} \\
\bottomrule
\end{tabular}}
\end{table*}

\subsection{Multiple organ reconstruction}
\label{supp:multiple_organ}

To test whether the approach generalizes beyond the lungs, we train a second model that predicts the heart and the aorta together with both lungs. The setup is otherwise identical to the lung model. Heart and aorta are taken from the TotalSegmentator~\cite{wasserthal2023totalsegmentator} masks. As the aorta is a tubular structure that may extend beyond the field of view, we cut the ground-truth aorta at the level of vertebra T10 for our model and all baselines.
For heart and aorta, the pose head was adapted to predict two additional scale factors, $k_{\text{heart}}$ and $k_{\text{aorta}}$ besides the lung scale $k$.

\begin{figure*}
    \centering
    \includegraphics[width=\textwidth]{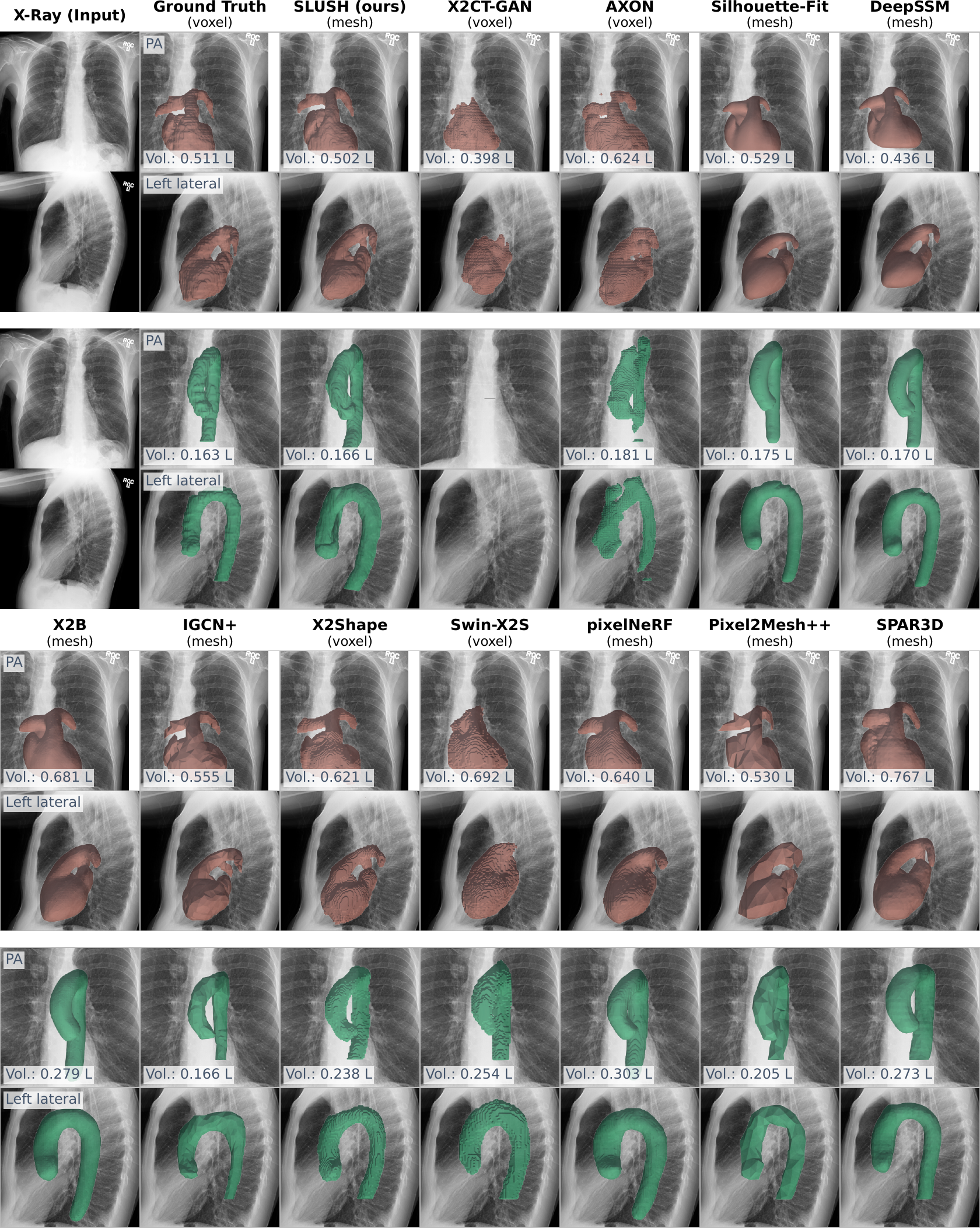}
    \caption{Qualitative comparison of a case based on a pair of bi-planar X-rays from the LIDC-RealPairs benchmark (LIDC-IDRI-0007). Predicted heart and aorta are overlaid on the input X-rays in the frame in which each method predicts them.}
    \label{fig:qual_heart_aorta_0007}
\end{figure*}

\begin{figure*}
    \centering
    \includegraphics[width=\textwidth]{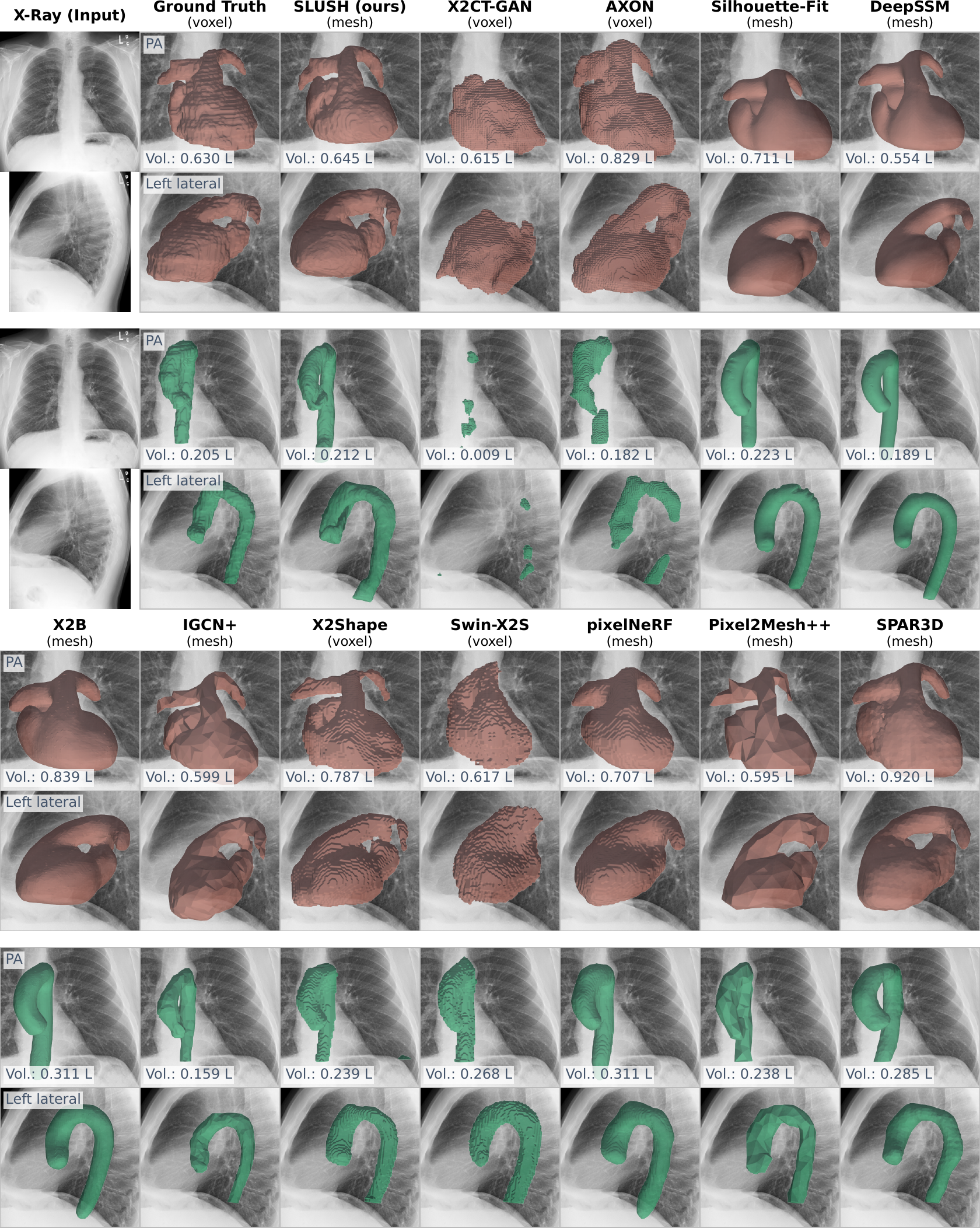}
    \caption{Qualitative comparison of a case based on a pair of bi-planar X-rays from the LIDC-RealPairs benchmark (LIDC-IDRI-0231). Predicted heart and aorta are overlaid on the input X-rays in the frame in which each method predicts them.}
    \label{fig:qual_heart_aorta_0231}
\end{figure*}

\Cref{fig:qual_heart_aorta_0007,fig:qual_heart_aorta_0231} show reconstructions for two LIDC-RealPairs cases, and \cref{tab:structures} reports the surface error (ASSD) and volume error (MAPE) per structure on the in-house test set and on LIDC-RealPairs. SLUSH performs best or on par for all structures on both datasets, with one exception: for the ASSD of the aorta it ranks second, behind DeepSSM~\cite{bhalodia2018deepssm} on the in-house test set and behind the SSM Silhouette-Fit on LIDC-RealPairs.

\begin{table*}
  \caption{Per-structure ASSD [mm] after rigid alignment (rotation + translation) and relative volume error MAPE [\%], mean over cases. Best value per column in bold, second best in italics. }
  \label{tab:structures}
  \centering
  \setlength{\tabcolsep}{3pt}
  \resizebox{\textwidth}{!}{%
  \begin{tabular}{@{}lcccccccccccccccc@{}}
    \toprule
     & \multicolumn{8}{c}{In-house test (n=200)} & \multicolumn{8}{c}{LIDC-RealPairs (n=179)} \\
    \cmidrule(lr){2-9} \cmidrule(lr){10-17}
     & \multicolumn{2}{c}{Left lung} & \multicolumn{2}{c}{Right lung} & \multicolumn{2}{c}{Heart} & \multicolumn{2}{c}{Aorta} & \multicolumn{2}{c}{Left lung} & \multicolumn{2}{c}{Right lung} & \multicolumn{2}{c}{Heart} & \multicolumn{2}{c}{Aorta} \\
    \cmidrule(lr){2-3} \cmidrule(lr){4-5} \cmidrule(lr){6-7} \cmidrule(lr){8-9} \cmidrule(lr){10-11} \cmidrule(lr){12-13} \cmidrule(lr){14-15} \cmidrule(lr){16-17}
    Method & ASSD [mm] $\downarrow$ & MAPE [\%] $\downarrow$ & ASSD [mm] $\downarrow$ & MAPE [\%] $\downarrow$ & ASSD [mm] $\downarrow$ & MAPE [\%] $\downarrow$ & ASSD [mm] $\downarrow$ & MAPE [\%] $\downarrow$ & ASSD [mm] $\downarrow$ & MAPE [\%] $\downarrow$ & ASSD [mm] $\downarrow$ & MAPE [\%] $\downarrow$ & ASSD [mm] $\downarrow$ & MAPE [\%] $\downarrow$ & ASSD [mm] $\downarrow$ & MAPE [\%] $\downarrow$ \\
    \midrule
    Mean Template & 8.40 & 34.05 & 8.22 & 27.33 & 4.09 & 21.36 & 4.07 & 28.44 & 8.13 & 27.38 & 8.38 & 28.08 & 4.05 & 26.56 & 3.85 & 29.46 \\
    X2CT-GAN~\cite{ying2019x2ct} & 7.80 & 21.72 & 13.43 & 48.39 & 9.28 & 24.15 & 22.59 & 80.09 & 7.87 & 22.27 & 14.29 & 53.02 & 8.80 & 27.46 & 22.78 & 81.08 \\
    AXON~\cite{rath2026axon} & 8.17 & 29.37 & 7.81 & 21.53 & 7.16 & 24.16 & 6.22 & 30.54 & 9.67 & 43.15 & 9.99 & 41.99 & 7.90 & 36.16 & 6.25 & 35.34 \\
    SSM Silhouette-Fit & 7.64 & 28.27 & 7.74 & 23.21 & \textit{3.91} & 17.54 & 3.45 & \textit{19.37} & 7.66 & 25.92 & 8.04 & 24.33 & 4.07 & 24.07 & \textbf{3.05} & \textit{16.73} \\
    DeepSSM~\cite{bhalodia2018deepssm} & 6.74 & 19.53 & 6.78 & 17.62 & \textbf{3.35} & \textit{13.22} & \textbf{3.31} & 19.75 & 6.79 & 18.42 & 6.65 & \textit{16.74} & \textit{3.23} & \textit{14.97} & 3.14 & 19.41 \\
    X2B~\cite{guven2026x2br} & 6.71 & 32.33 & \textit{5.81} & 17.13 & 5.14 & 30.60 & 3.82 & 48.76 & 6.89 & 33.85 & 6.55 & 28.85 & 5.48 & 40.88 & 3.98 & 58.76 \\
    IGCN+\cite{nakao2022igcn} & 7.09 & 22.31 & 7.16 & 20.58 & 4.47 & 17.48 & 4.38 & 28.69 & 6.83 & 20.63 & 7.28 & 21.09 & 4.18 & 16.91 & 4.07 & 23.93 \\
    X2Shape~\cite{PAN2026104085} & 6.87 & 20.18 & 6.96 & 17.81 & 5.58 & 22.29 & 3.57 & 21.09 & 7.08 & 21.82 & 7.25 & 19.32 & 5.49 & 28.16 & 3.56 & 23.75 \\
    Swin-X2S~\cite{liu2025swin} & 6.93 & 23.00 & 6.73 & 19.65 & 6.16 & 27.13 & 4.38 & 42.79 & 6.93 & 22.65 & 6.83 & 21.44 & 6.65 & 43.70 & 4.66 & 54.21 \\
    pixelNeRF~\cite{yu2021pixelnerf} & 6.84 & 30.71 & 6.49 & 20.26 & 4.68 & 22.00 & 4.53 & 69.51 & 7.26 & 36.95 & 6.72 & 25.01 & 5.02 & 32.47 & 4.98 & 82.54 \\
    Pixel2Mesh++~\cite{wen2019pixel2mesh++} & \textit{6.09} & \textit{16.90} & 6.23 & \textit{15.29} & 4.62 & 16.73 & 4.04 & 27.92 & \textit{6.35} & 18.90 & \textit{6.48} & 18.59 & 4.94 & 29.26 & 3.96 & 35.83 \\
    SPAR3D~\cite{huang2025spar3d} & 6.96 & 21.30 & 7.16 & 20.91 & 5.36 & 38.92 & 4.61 & 32.00 & 6.88 & \textit{18.08} & 7.11 & 21.93 & 6.03 & 48.35 & 4.57 & 35.59 \\
    \textbf{SLUSH (ours)} & \textbf{4.84} & \textbf{15.50} & \textbf{4.94} & \textbf{14.75} & \textbf{3.35} & \textbf{12.22} & \textit{3.32} & \textbf{14.87} & \textbf{4.47} & \textbf{14.06} & \textbf{4.42} & \textbf{13.20} & \textbf{3.12} & \textbf{11.97} & \textit{3.13} & \textbf{13.97} \\
    \bottomrule
  \end{tabular}}
\end{table*}

\subsection{Nodule localization setup}
\label{supp:nodule_localization}

The 2D-to-3D nodule localization experiment in \cref{tab:nodule_backprojection_template_icp_hull} is set up as follows:

\begin{enumerate}[leftmargin=*, itemsep=2pt, topsep=4pt]
  \item \textbf{Nodule selection.} LIDC-IDRI provides radiologist segmentations of lung nodules in the CT. As small nodules are hardly visible on radiographs, we select 26 nodules that have a diameter of more than 20\,mm, are visible on both X-ray views and were segmented by at least two annotators. The consensus mask is obtained by voxel-wise majority vote over the annotators.
  \item \textbf{2D annotations.} The CT mask of each nodule is projected onto both X-rays using the xvr reference cameras, which yields one nodule mask per view. The CT is used only to create these masks and to evaluate the result.
  \item \textbf{Back-projection.} Each mask is back-projected from the detector towards the source, using the intrinsics of the X-ray and the extrinsics predicted by the pose head (\cref{fig:teaser}). The nodule is localized in 3D where the rays of both views intersect.
  \item \textbf{Baselines.} The baselines correct no extrinsics and use the template PA and lateral cameras. As unregistered X-rays would often prevent the two projection cones from meeting at all, each X-ray is first aligned rigidly in the image plane (rotation and translation) to a template DRR of the same view.
  \item \textbf{Evaluation.} The nodule estimation is compared with the annotated CT nodule after rigid ICP alignment of the predicted lungs to the reference lungs. A nodule counts as a hit if the estimate overlaps the CT nodule (Dice $>0$).
\end{enumerate}

\paragraph{Control with ground-truth lungs.}
To separate the effect of the cameras from that of the shape, we repeat the experiment with the ground-truth lungs and the template cameras. Even with perfect shapes, only 8 of 26 nodules are hit. Localization therefore depends mainly on correct extrinsics, which explains the advantage of SLUSH.

\subsection{Additional qualitative results}
\label{supp:additional_results}

\begin{figure*}
    \centering
    \includegraphics[width=\textwidth]{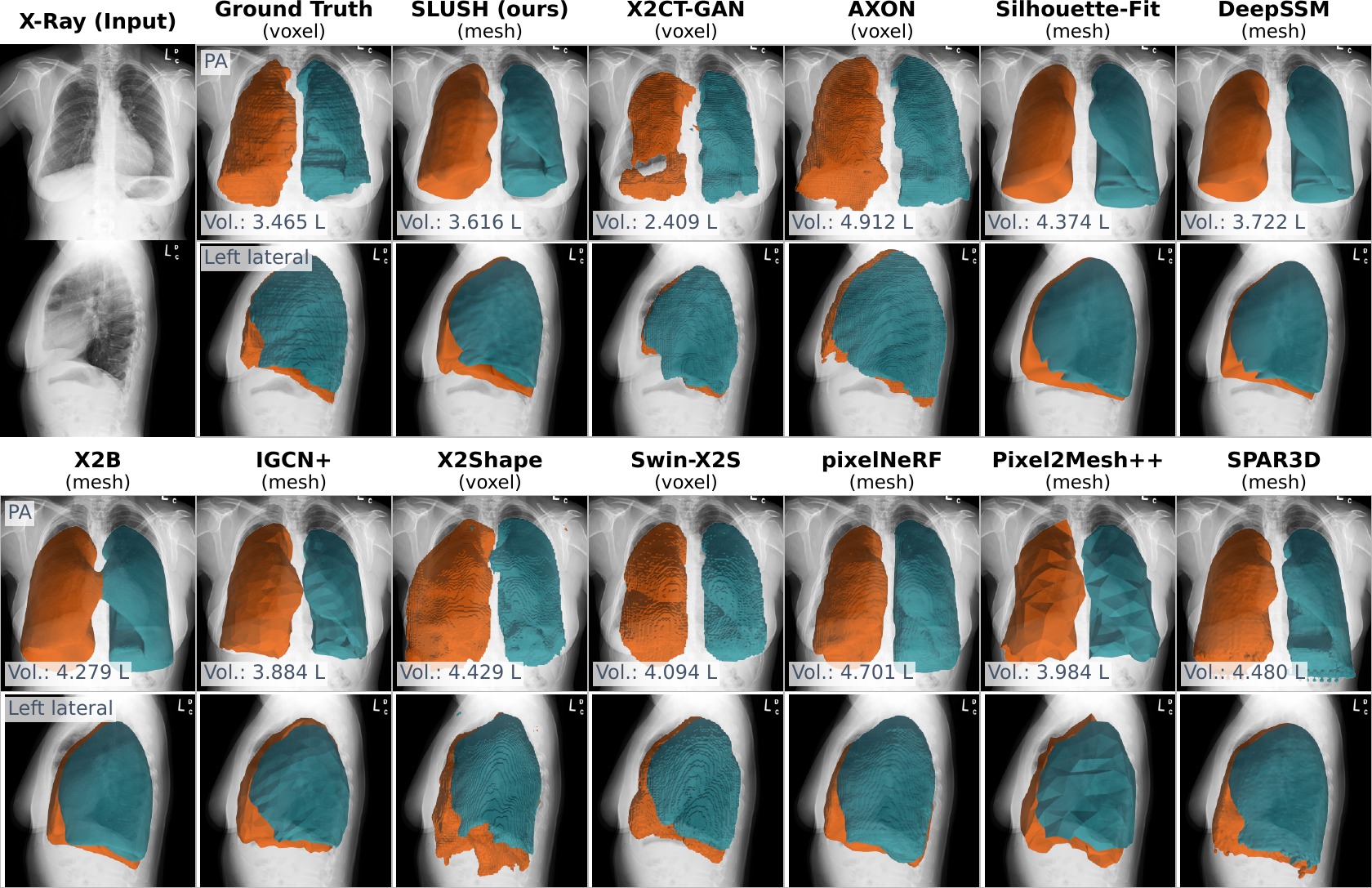}
    \caption{Qualitative comparison of a case based on a pair of bi-planar X-rays from the LIDC-RealPairs benchmark (LIDC-IDRI-0276). Predicted right (orange) and left (teal) lungs are overlaid on the input X-rays in the frame in which each method predicts them.}
    \label{fig:qual_0276}
\end{figure*}

\begin{figure*}
    \centering
    \includegraphics[width=\textwidth]{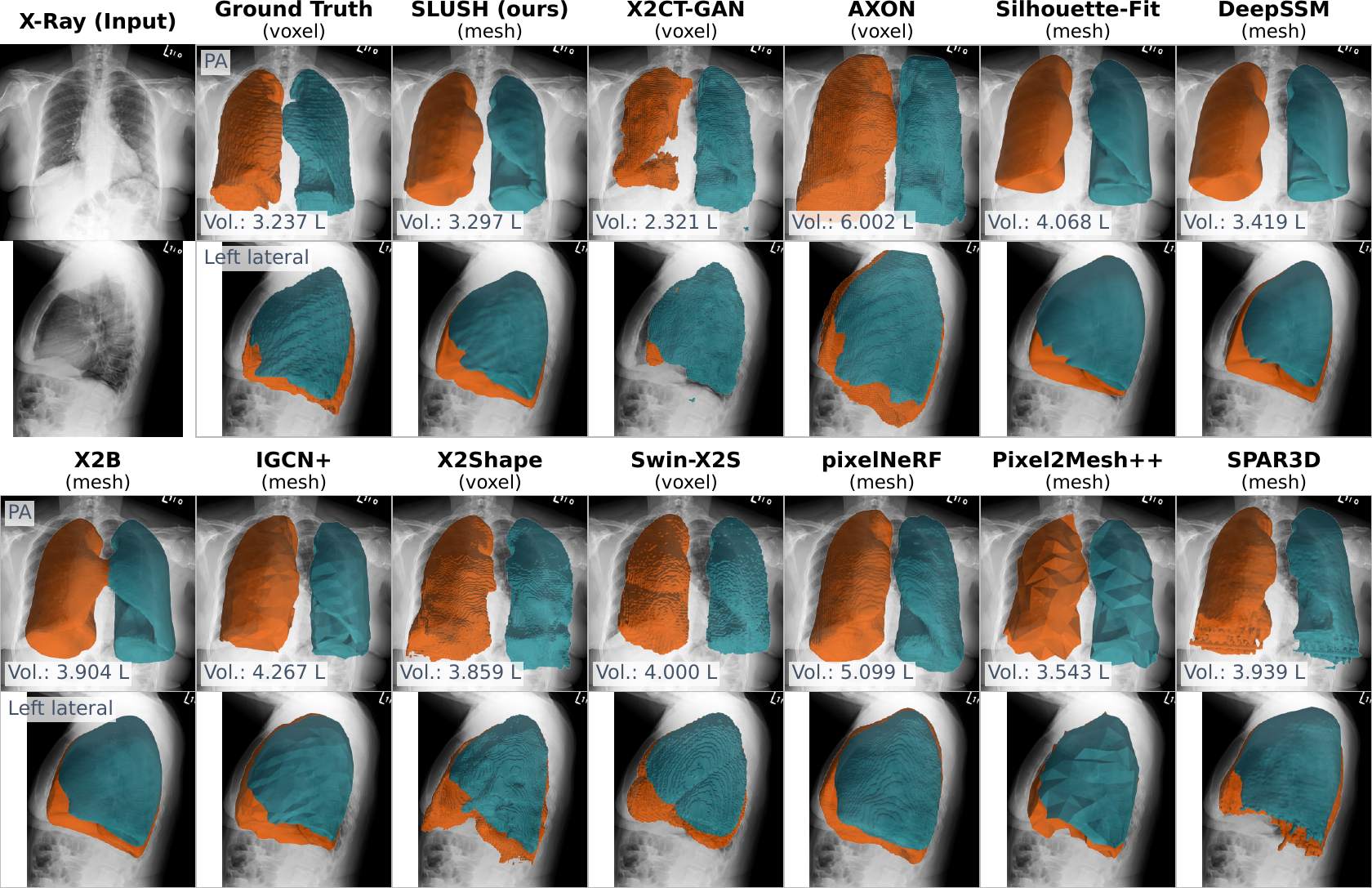}
    \caption{Qualitative comparison of a case based on a pair of bi-planar X-rays from the LIDC-RealPairs benchmark (LIDC-IDRI-0139). Predicted right (orange) and left (teal) lungs are overlaid on the input X-rays in the frame in which each method predicts them.}
    \label{fig:qual_0139}
\end{figure*}

\begin{figure*}
    \centering
    \includegraphics[width=\textwidth]{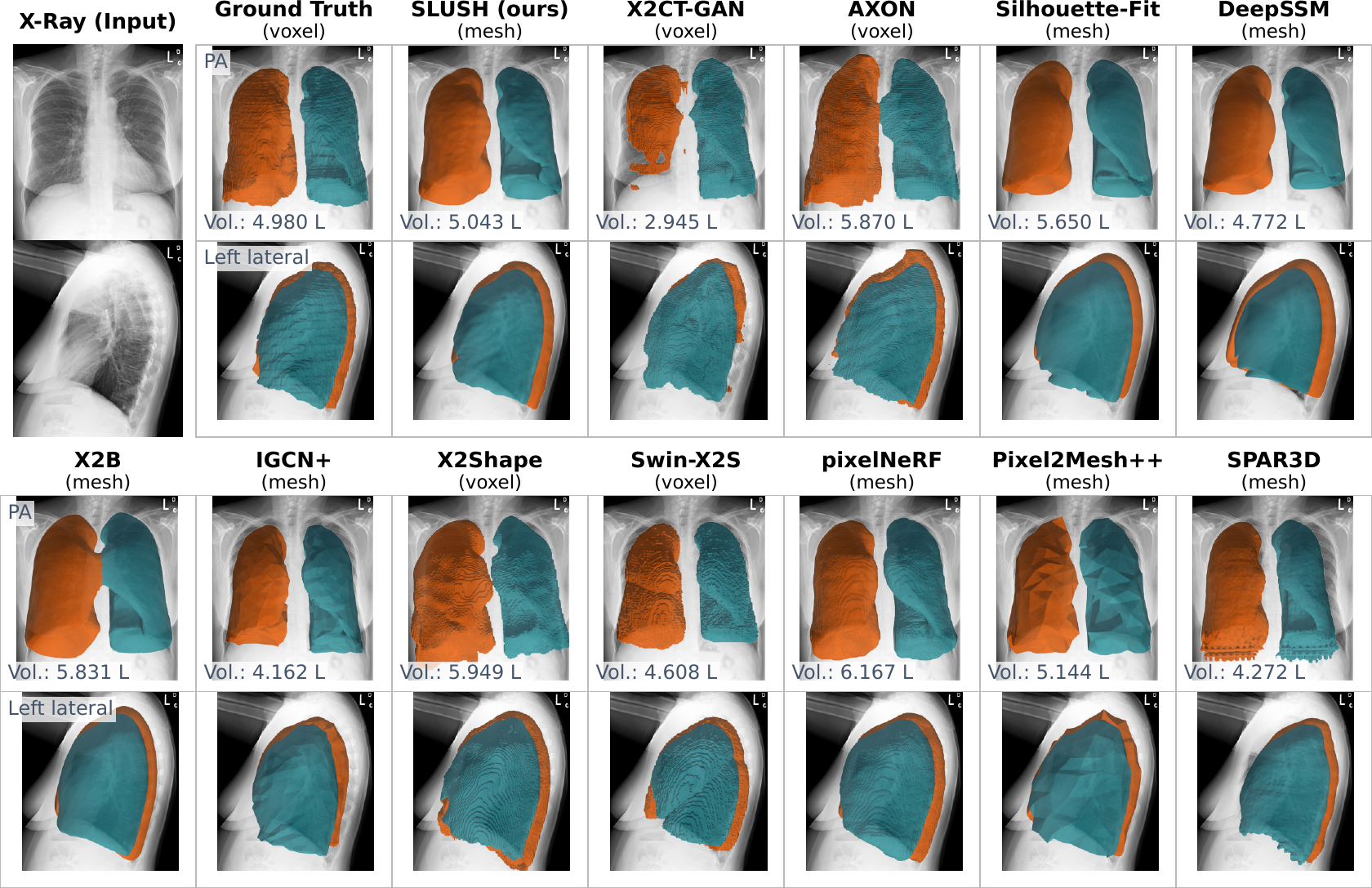}
    \caption{Qualitative comparison of a case based on a pair of bi-planar X-rays from the LIDC-RealPairs benchmark (LIDC-IDRI-0144). Predicted right (orange) and left (teal) lungs are overlaid on the input X-rays in the frame in which each method predicts them.}
    \label{fig:qual_0144}
\end{figure*}

\begin{figure*}
    \centering
    \includegraphics[width=\textwidth]{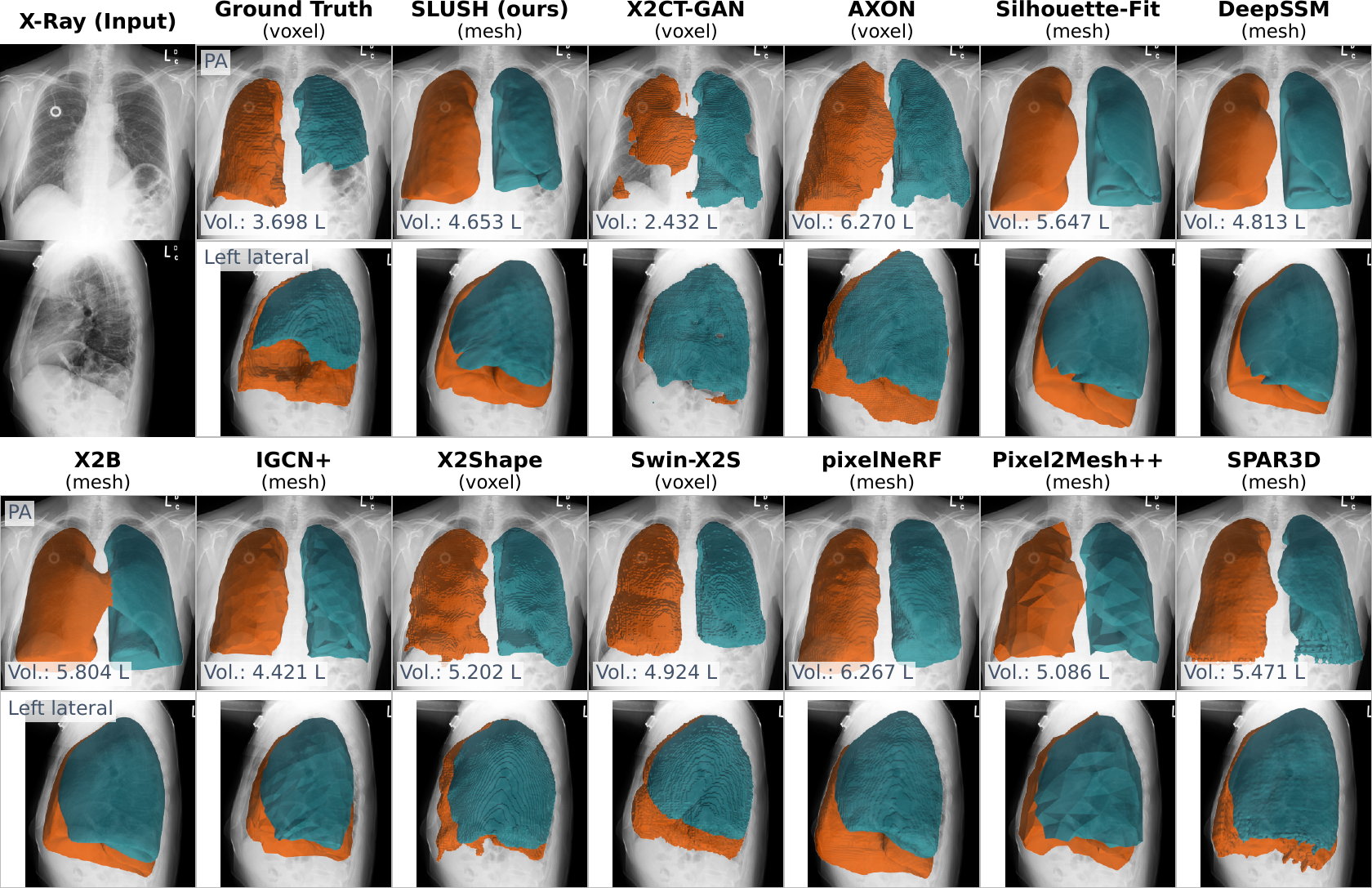}
    \caption{Qualitative comparison of a failure case based on a pair of bi-planar X-rays from the LIDC-RealPairs benchmark (LIDC-IDRI-0033). Predicted right (orange) and left (teal) lungs are overlaid on the input X-rays in the frame in which each method predicts them.}
    \label{fig:qual_0033_fail}
\end{figure*}

In addition to the sample of results presented in the main paper, we show further qualitative results. \cref{fig:qual_0276} and \cref{fig:qual_0139} show a comparison of smaller lungs than in the main paper, whereas \cref{fig:qual_0144} shows a larger lung. \Cref{fig:qual_0033_fail} shows a failure case, where both our model and the baselines fail to accurately predict the left lung. It seems that the inflation of the PA view differs slightly from that of the lung shown in the lateral X-ray and the ground truth. Since our model seems to follow the PA X-ray, it does not accurately reconstruct the lung as expected.

\subsection{Pose head in other architectures}
\label{supp:baseline_posehead}

To test whether the benefit of the SLUSH pose head is specific to our architecture, we add it to two baselines and train it jointly with each of them. Pixel2Mesh++~\cite{wen2019pixel2mesh++} is the strongest baseline, and pixelNeRF~\cite{yu2021pixelnerf} samples image features in 3D in the same way as our back-projection.

Both receive the same pose head as SLUSH, a ResNet-18 on both views together with the $2\times28$ mask statistics. It predicts a 6-DoF pose per view, is trained with the same pose loss, and its cameras replace the fixed canonical cameras. Pixel2Mesh++ uses the predicted cameras in its perceptual feature pooling, which projects the mesh vertices into both views. To make this differentiable with respect to the pose, we replace its nearest-pixel lookup with bilinear sampling. pixelNeRF samples pixel-aligned image features at 3D query points. The predicted cameras enter this lookup, and the network is otherwise unchanged.

\Cref{tab:whole_lung_shape_volume_jointpose} shows that both baselines improve on both datasets. The ASSD decreases by 7\,\% (in-house) and 12\,\% (LIDC-RealPairs) for Pixel2Mesh++ and by 24\,\% and 28\,\% for pixelNeRF. As for SLUSH, the gain is larger on LIDC-RealPairs. This indicates that the acquisition geometry is an important limitation on real X-rays across different reconstruction paradigms and that the pose head is a general component.

\begin{table*}[t]
  \caption{Shape and volume error (whole lung) of the original baselines and their variants trained with our pose head. Shape metrics after rigid alignment (rotation + translation); mean $\pm$ SD over cases. Best value per column in bold, second best in italics.}
  \label{tab:whole_lung_shape_volume_jointpose}
  \centering
  \setlength{\tabcolsep}{3pt}
  \resizebox{\textwidth}{!}{%
  \begin{tabular}{@{}lcccccccccccc@{}}
    \toprule
     & \multicolumn{6}{c}{In-house test (n=200)} & \multicolumn{6}{c}{LIDC-RealPairs (n=179)} \\
    \cmidrule(lr){2-7} \cmidrule(lr){8-13}
    Method & ASSD [mm] $\downarrow$ & HD95 [mm] $\downarrow$ & MeshDice $\uparrow$ & MAE [L] $\downarrow$ & MAPE [\%] $\downarrow$ & Pearson r $\uparrow$ & ASSD [mm] $\downarrow$ & HD95 [mm] $\downarrow$ & MeshDice $\uparrow$ & MAE [L] $\downarrow$ & MAPE [\%] $\downarrow$ & Pearson r $\uparrow$ \\
    \midrule
    Pixel2Mesh++~\cite{wen2019pixel2mesh++} & 6.14\,{\scriptsize$\pm$1.56} & 18.18\,{\scriptsize$\pm$5.61} & 0.829\,{\scriptsize$\pm$0.050} & 0.601\,{\scriptsize$\pm$0.455} & \textit{14.40}\,{\scriptsize$\pm$12.66} & 0.828 & 6.37\,{\scriptsize$\pm$1.92} & 17.66\,{\scriptsize$\pm$6.04} & 0.825\,{\scriptsize$\pm$0.058} & 0.679\,{\scriptsize$\pm$0.538} & 16.80\,{\scriptsize$\pm$17.65} & 0.757 \\
    Pixel2Mesh++~\cite{wen2019pixel2mesh++} + pose head & 5.73\,{\scriptsize$\pm$1.63} & 16.52\,{\scriptsize$\pm$5.12} & 0.842\,{\scriptsize$\pm$0.052} & \textit{0.589}\,{\scriptsize$\pm$0.485} & 14.82\,{\scriptsize$\pm$15.29} & 0.817 & 5.61\,{\scriptsize$\pm$1.65} & 15.53\,{\scriptsize$\pm$5.00} & 0.848\,{\scriptsize$\pm$0.051} & \textit{0.612}\,{\scriptsize$\pm$0.473} & \textit{15.40}\,{\scriptsize$\pm$14.93} & 0.821 \\
    \midrule
    pixelNeRF~\cite{yu2021pixelnerf} & 6.61\,{\scriptsize$\pm$1.93} & 19.71\,{\scriptsize$\pm$4.74} & 0.832\,{\scriptsize$\pm$0.059} & 0.834\,{\scriptsize$\pm$0.645} & 22.95\,{\scriptsize$\pm$22.62} & 0.816 & 6.90\,{\scriptsize$\pm$2.02} & 19.49\,{\scriptsize$\pm$4.70} & 0.828\,{\scriptsize$\pm$0.061} & 1.030\,{\scriptsize$\pm$0.680} & 27.29\,{\scriptsize$\pm$23.24} & 0.816 \\
    pixelNeRF~\cite{yu2021pixelnerf} + pose head & \textit{5.02}\,{\scriptsize$\pm$1.95} & \textbf{13.86}\,{\scriptsize$\pm$5.45} & \textit{0.864}\,{\scriptsize$\pm$0.061} & 0.624\,{\scriptsize$\pm$0.529} & 17.56\,{\scriptsize$\pm$19.30} & \textbf{0.860} & \textit{4.97}\,{\scriptsize$\pm$1.89} & \textit{12.94}\,{\scriptsize$\pm$4.90} & \textit{0.868}\,{\scriptsize$\pm$0.057} & 0.736\,{\scriptsize$\pm$0.526} & 19.46\,{\scriptsize$\pm$18.01} & \textit{0.865} \\
    \midrule
    \textbf{SLUSH (ours)} & \textbf{4.87}\,{\scriptsize$\pm$1.60} & \textit{14.62}\,{\scriptsize$\pm$5.70} & \textbf{0.869}\,{\scriptsize$\pm$0.049} & \textbf{0.554}\,{\scriptsize$\pm$0.435} & \textbf{14.04}\,{\scriptsize$\pm$14.48} & \textit{0.850} & \textbf{4.42}\,{\scriptsize$\pm$1.30} & \textbf{12.58}\,{\scriptsize$\pm$3.60} & \textbf{0.882}\,{\scriptsize$\pm$0.041} & \textbf{0.474}\,{\scriptsize$\pm$0.385} & \textbf{11.66}\,{\scriptsize$\pm$11.73} & \textbf{0.891} \\
    \bottomrule
  \end{tabular}}
\end{table*}

SLUSH remains the best method on all metrics on LIDC-RealPairs and on most metrics on the in-house test set, where pixelNeRF with pose head reaches a lower HD95 and a higher volume correlation. Unlike pixelNeRF, SLUSH predicts meshes with vertex correspondence directly and estimates the volume more accurately (MAPE 14.0\,\% vs.\ 17.6\,\% in-house and 11.7\,\% vs.\ 19.5\,\% on LIDC-RealPairs).

\end{document}